\documentclass[preprint, 12pt]{elsarticle}

\usepackage{lipsum}
\usepackage[utf8]{inputenc}
\usepackage[margin=1in]{geometry}
\usepackage{bold-extra}
\usepackage{xcolor}
\usepackage{adjustbox}
\usepackage{diagbox}
\usepackage{appendix}
\usepackage{lineno}

\usepackage{amssymb}
\usepackage{amsthm}
\usepackage{amsmath}
\usepackage{algorithm}
\usepackage{algpseudocode}
\usepackage{graphicx}
\usepackage{times}
\usepackage[titles]{tocloft}
\usepackage{enumerate}
\usepackage{bm}
\usepackage{sidecap}
\usepackage{epstopdf}
\usepackage{pdfpages}
\usepackage[colorinlistoftodos]{todonotes}
\usepackage{soul}
\usepackage{multicol}
\usepackage{float}
\usepackage{multibib}
\usepackage{geometry}
\usepackage[small,compact]{titlesec}
\usepackage[normal,bf,labelsep=colon,skip=2pt]{caption}
\usepackage{setspace}
\usepackage{subfig}
\usepackage{tocloft}
\usepackage{url}
\usepackage{floatflt}
\usepackage{array}
\usepackage{pgfgantt}
\usepackage{sidecap}
\usepackage{cite}
\usepackage{multirow}
\usepackage{arydshln}
\usepackage[hidelinks]{hyperref}

\numberwithin{equation}{section}
\numberwithin{figure}{section}
\numberwithin{table}{section}

\DeclareMathOperator*{\argmin}{arg\,min}

\begin{document}

\begin{frontmatter}

\title{Modeling Partially Observed Nonlinear Dynamical Systems and Efficient Data Assimilation via Discrete-Time Conditional Gaussian Koopman Network}
\author[1]{Chuanqi Chen}
\ead{cchen656@wisc.edu}
\author[2]{Zhongrui Wang}
\ead{wang3262@wisc.edu}
\author[2]{Nan Chen}
\ead{chennan@math.wisc.edu}
\author[1]{Jin-Long Wu\corref{cor1}} \ead{jinlong.wu@wisc.edu}
\cortext[cor1]{Corresponding author}

\address[1]{Department of Mechanical Engineering, University of Wisconsin–Madison, Madison, WI 53706}
\address[2]{Department of Mathematics, University of Wisconsin–Madison, Madison, WI 53706}

\begin{abstract}
A discrete-time conditional Gaussian Koopman network (CGKN) is developed in this work to learn surrogate models that can perform efficient state forecast and data assimilation (DA) for high-dimensional complex dynamical systems, e.g., systems governed by nonlinear partial differential equations (PDEs). Focusing on nonlinear partially observed systems that are common in many engineering and earth science applications, this work exploits Koopman embedding to discover a proper latent representation of the unobserved system states, such that the dynamics of the latent states are conditional linear, i.e., linear with the given observed system states. The modeled system of the observed and latent states then becomes a conditional Gaussian system, for which the posterior distribution of the latent states is Gaussian and can be efficiently evaluated via analytical formulae. The analytical formulae of DA facilitate the incorporation of DA performance into the learning process of the modeled system, which leads to a framework that unifies scientific machine learning (SciML) and data assimilation. The performance of discrete-time CGKN is demonstrated on several canonical problems governed by nonlinear PDEs with intermittency and turbulent features, including the viscous Burgers' equation, the Kuramoto-Sivashinsky equation, and the 2-D Navier-Stokes equations, with which we show that the discrete-time CGKN framework achieves comparable performance as the state-of-the-art SciML methods in state forecast and provides efficient and accurate DA results. The discrete-time CGKN framework also serves as an example to illustrate unifying the development of SciML models and their other outer-loop applications such as design optimization, inverse problems, and optimal control.
\end{abstract}

\begin{keyword}
Scientific machine learning \sep
Data assimilation \sep
Spatiotemporal dynamical systems \sep 
Computational mechanics \sep
Uncertainty quantification \sep
Koopman theory
\end{keyword}

\end{frontmatter}

\clearpage

\section{Introduction}
\label{sec:Introduction}

Complex dynamical systems are ubiquitous across a wide range of science and engineering disciplines, including solid mechanics, fluid dynamics, geophysics, neuroscience, and material science \citep{jost2005dynamical, wiggins2003introduction, majda2006nonlinear, chen2023stochastic, vogiatzoglou2025physics, wang2025development}. Many of those systems are described by partial differential equations (PDEs) or their stochastic extensions and characterized by complex features such as nonlinearity, multiscale dynamics, chaos, turbulence, shock behavior, intermittency, and non-Gaussian statistics \citep{dijkstra2013nonlinear, trenberth2015attribution, moffatt2021extreme, majda2003introduction, majda2018model,manneville1979intermittency}. Deriving governing equations for these spatiotemporal dynamical systems from first principles is a conventional way of modeling and understanding these systems. The advantages of having explicit governing equations include adherence to physical laws and interpretability. However, deriving the governing equations usually requires a deep understanding of the underlying physics, extensive domain knowledge, and strong prior assumptions. Additionally, computationally expensive numerical simulation is typically necessary to understand and predict most of these systems, which can further increase the burden of downstream applications such as data assimilation (DA), uncertainty quantification (UQ), and decision-making. As a result, data-driven methods, including closure models \citep{sanderse2024scientific, wu2024learning, chen2024physics}, sparse identification \citep{rudy2017data, schaeffer2013sparse}, reduced order models \citep{majda2018strategies, carlberg2013gnat, noack2011reduced}, causality-based models \citep{chen2023causality, chen2023ceboosting, chen2024minimum}, and Gaussian processes \citep{chen2021solving} have become increasingly promising and practical for studying spatiotemporal dynamical systems and carrying out many of the tasks mentioned above.

As one of these data-driven methods, neural-network-based models in deep learning \citep{lecun2015deep, krizhevsky2012imagenet, rumelhart1986learning, lecun1995convolutional, lecun1989handwritten} have promoted the field of scientific machine learning (SciML) and have shown promise in the study of spatiotemporal dynamical systems with the rapid growth of available data and computing resources \citep{karniadakis2021physics, duraisamy2019turbulence, yu2024learning, wang2017physics, wu2018physics}. Based on the universal approximation theorem \citep{cybenko1989approximation, hornik1989multilayer, chen1995universal}, neural networks can approximate any nonlinear functions and operators. Consequently, they demonstrate significant capability in discovering unknown dynamics and making predictions. For example, as a continuous-in-time version of ResNet \citep{he2016deep}, the neural ODE \citep{chen2018neural} is proposed to learn the dynamics of systems with efficient memory usage. Physics-informed neural networks \citep{raissi2019physics}, which integrate physical laws into the neural networks, are developed to solve forward and inverse PDE-governed problems. Designed to learn mappings between infinite-dimensional function spaces, neural operators \citep{lu2021learning, li2021fourier, chen2025neural, zhang2024modno,lin2023b, wu2023comprehensive} have been widely used to approximate solution operators and build spatiotemporal models. Recently, generative models \citep{kingma2013auto,goodfellow2020generative, wu2020enforcing, ho2020denoising, song2020score} have been utilized to construct stochastic nonlocal models \citep{dong2025data}, generate spatiotemporal turbulence \citep{du2024conditional}, capture extreme event statistics \citep{stamatelopoulos2025can} and learn PDE solutions \citep{gao2025generative}. Additionally, active learning \citep{settles2009active, konyushkova2017learning} has been considered for identifying informative data \citep{huan2013simulation, von2011bayesian}, simulating nonlinear systems \citep{kapadia2024active}, estimating model parameters \citep{atkinson2007optimum}, and learning model discrepancies \citep{yang2025active}. 

In recent years, leveraging online observational data to improve the performance of data-driven models has become a critical component of reliably deploying those models in real-world applications, which motivates the joint study of SciML and DA. DA integrates observational data with computational models to enhance state estimation \citep{kalman1960new, kalman1961new, majda2012filtering, law2015data}. It is extensively applied in real-time forecasting, parameter estimation, imputation, and optimal control \citep{schneider2021learning, schneider2022ensemble, iglesias2013ensemble, chen2021efficient, chen2018efficient}. The governing equations of the underlying dynamics are utilized as computational models for short-term statistical prediction, known as the prior distribution. Through Bayesian inference, the prior distribution is updated with the observational data to form the so-called posterior distribution, which is the solution of DA. To handle general nonlinear dynamical systems, the ensemble DA methods \citep{evensen1994sequential, bergemann2012ensemble, evensen2003ensemble, whitaker2002ensemble, burgers1998analysis} that utilize samples to approximate probabilistic distributions have become commonly used techniques. However, the direct application of ensemble DA methods to high-dimensional spatiotemporal systems is often computationally intractable, primarily due to the substantial computational cost incurred by the curse of dimensionality in sampling processes and the numerical simulations of PDEs. Although the adoption of efficient computational models, such as data-driven surrogate models \citep{mou2021data, peherstorfer2015dynamic, hijazi2020data, lin2021data}, reduced order models \citep{xiao2024operator, mou2023combining}, and stochastic parameterizations \citep{berner2017stochastic, mana2014toward, dawson2015simulating, chen2024stochastic}, coupled with various DA techniques like localization \citep{Anderson2001EAKF, malartic2022state} and empirical tunings of ensemble sizes and covariances matrix, can mitigate costs and facilitate the implementation of DA, the DA process still demands significant computational resources, and may suffer from sampling and approximation errors. Recently, deep learning has been employed to support DA for spatiotemporal systems \citep{cheng2023machine}, assisting the construction of computational models by neural networks that enhance forecast capability within DA \citep{rasp2018deep, bonavita2020machine, brajard2021combining}. Additionally, statistical variational DA methods have been proposed to generate observations in scenarios where observational data is scarce \citep{benaceur2024statistical}. Using encoder-decoder networks in latent DA has proven effective in utilizing multi-domain data for high-dimensional systems \citep{cheng2024multi}. Deep learning can also facilitate using data-driven ensembles at a low computational cost to accelerate ensemble DA \citep{chattopadhyay2023deep}. Furthermore, end-to-end deep learning methods have been developed to streamline the entire DA process \citep{revach2022kalmannet, boudier2023data}. 

While these existing methods have facilitated the implementation of DA with affordable computational costs and demonstrated successes in many applications, efficient DA methods for spatiotemporal dynamical systems still require further exploration and development, especially for incorporating DA into the training of SciML models to enhance their performance. Several recent works have started to explore how to incorporate DA into the training of a machine learning model from various perspectives, e.g., using ensemble Kalman filter to enhance the learning of structural errors for non-ergodic systems via derivative-free optimization~\citep{wu2024learning}, developing auto-differentiable ensemble Kalman filter~\citep{chen2022autodifferentiable,chen2023reduced}, exploiting analytical formulae of DA for a conditional Gaussian neural stochastic differential equation~\citep{chen2024cgnsde}. Importantly, developing a unified modeling framework that can handle both efficient state forecast and DA for spatiotemporal dynamical systems, with a low training cost and robust testing results for problems with high-dimensional (or even infinite-dimensional) system states, can significantly strengthen the joint study of SciML models and their data assimilation performance, potentially advancing the reliable deployment of SciML models in real-world applications.

To establish a unified deep learning framework for SciML modeling and efficient DA, the continuous-time conditional Gaussian Koopman network (CGKN) was proposed in~\citep{chen2025cgkn}, aiming to perform efficient state forecast and DA for the modeling and simulation of spatiotemporal dynamical systems. In contrast to the standard usage of Koopman theory \citep{koopman1931hamiltonian}, which fully linearizes nonlinear dynamical systems \citep{budivsic2012applied, mezic2013analysis, williams2016kernel, pan2020physics, lusch2018deep, dietrich2020koopman, lu2020prediction,otto2021koopman, bevanda2021koopman,lu2021extended,brunton2022modern}, CGKN leverages the Koopman embedding to discover a proper latent representation for the unobserved system states and employs a conditional linear structure to construct the data-driven model architecture. The resulting system is still highly nonlinear. Nevertheless, it leads to a conditional Gaussian system \citep{chen2018conditional, chen2022conditional, chen2016filtering, chen2024cgnsde, chen2025cgkn}, which encompasses numerous nonlinear models across various disciplines with wide-ranging applications in natural science and engineering \citep{chen2014information, chen2015noisy, chen2017beating, wang2024closed}. One desirable feature of the conditional Gaussian system is that it allows analytical DA formulae and thus avoids the need for ensemble DA methods. Together with a lower-dimensional latent representation for unobserved states, significantly improved efficiency can be achieved for the DA process in both the training and testing phases. In this work, we establish a discrete-time CGKN framework, which avoids numerical integration with small step sizes for the continuous CGKN framework in~\citep{chen2025cgkn} and thus significantly enhances the efficiency of handling problems with high-dimensional or even infinite-dimensional state variables. The effectiveness and efficiency of the discrete-time CGKN framework are demonstrated through several PDE-governed problems, including the viscous Burgers' equation, the Kuramoto–Sivashinsky equation, and the 2-D Navier-Stokes equations. 

The rest of this paper is organized as follows. Section~\ref{sec:Methodology} presents the discrete-time CGKN framework. Three numerical examples are presented in Section~\ref{sec:NumericalExperiments} to demonstrate the performance of the proposed framework. Section~\ref{sec:Conclusion} concludes the paper.

\section{Methodology}
\label{sec:Methodology}

Considering a spatiotemporal dynamical system including partial differential equations (PDEs) of the general form
\begin{equation}
\label{eq:CTCS}
    \partial_t \mathbf{u} = \mathcal{M}(\mathbf{u}),
\end{equation}
where $\mathbf{u}(\mathbf{x}, t) \in \mathbb{R}^{d_\mathbf{u}}$ is the system state, with spatial variable $\mathbf{x}\in D_{\mathbf{x}} \subset \mathbb{R}^{d_\mathbf{x}}$ and temporal variable $t \in [0, L_t]$, and $\mathcal{M}$ is a nonlinear operator. When discretized in both spatial and temporal domains, Eq.~\eqref{eq:CTCS} can be written as a discrete dynamical system:
\begin{equation}
\label{eq:DTDS}
    \mathbf{u}^{n+1} = \mathcal{G}\big( \mathbf{u}^{n} \big),
\end{equation}
where $\mathbf{u}^{n} := \mathbf{u}(\cdot, t_n)|_{D_{\mathbf{x}}^{(M)}} \in \mathbb{R}^{d}$ denotes the system state variables evaluated at time $t_n$ and $M$-point spatial discretization $D_{\mathbf{x}}^{(M)}=\{\mathbf{x}^{(1)}, \mathbf{x}^{(2)}, \cdots, \mathbf{x}^{(M)}\}$ of the bounded domain $D_{\mathbf{x}}$. $\mathcal{G}$ is the solution map for this discrete system over time interval $t_{n+1} - t_n = \Delta t$. This map $\mathcal{G}$ is also referred to as the state transition map. It should be noted that the temporal resolution of the data, dictated by the choice of $\Delta t$, is often larger than the numerical integration time step $\delta t$, which is required by the numerical simulation of Eq.~\eqref{eq:CTCS} and is typically small to ensure accuracy and stability of the simulation. Therefore, learning and applying the solution map $\mathcal{G}$ at a coarser temporal resolution $\Delta t$ can significantly reduce the computational cost.

In this work, we focus on the situation with partial observations, which means the online observational data only contain a subset of the system states of the discrete dynamical system~\eqref{eq:DTDS}. The observed states and unobserved states are denoted as $\mathbf{u}_1^n \in \mathbb{R}^{d_1}$ and $\mathbf{u}_2^n \in \mathbb{R}^{d_2}$, respectively, with $\mathbf{u}^n = \{\mathbf{u}_1^n, \mathbf{u}_2^n\}$. For instance, $\mathbf{u}_1^n$ and $\mathbf{u}_2^n$ can be interpreted as the spatial partition of $\mathbf{u}^n$. Consequently, the system in Eq.~\eqref{eq:DTDS} can be rewritten as follows:
\begin{equation}
\label{eq:DTDS_u1u2}
\begin{aligned}
    \mathbf{u}_1^{n+1} &= \mathcal{G}_1\big( \mathbf{u}_1^{n}, \mathbf{u}_2^{n}\big), \\
    \mathbf{u}_2^{n+1} &= \mathcal{G}_2\big( \mathbf{u}_1^{n}, \mathbf{u}_2^{n}\big),
\end{aligned}
\end{equation}
where the two maps $\mathcal{G}_1$ and $\mathcal{G}_2$ are often both nonlinear for many real-world applications. These nonlinear maps pose challenges to data assimilation (DA), e.g., preventing the direct use of the analytical formulae and thus demanding methods developed for general nonlinear problems such as ensemble Kalman filter or even particle filters that involve the evaluation of high-dimensional quadratures~\citep{cheng2024efficient}.

The goal of this work is to learn a surrogate model that embeds a conditional Gaussian structure, which is capable of performing two tasks for systems governed by Eq.~\eqref{eq:DTDS_u1u2}: efficient state forecast and DA. The task of DA aims to the conditional distribution $p(\mathbf{u}_2^{n} | \{\mathbf{u}_1^{i}\}_{i=0}^{n})$, while the task of state forecast is to predict the future states $\mathbf{u}^{n}$ without assuming the knowledge for the exact initial unobserved states $\mathbf{u}_2^{0}$. 

More specifically, to achieve efficient DA via the analytical formulae from the conditional Gaussian structure, we choose to avoid directly dealing with the nonlinearity of the unobserved state $\mathbf{u}_2$. According to the Koopman theory~\citep{koopman1931hamiltonian}, a nonlinear dynamical system can be linearized by modeling the evolution the observation function of the system states, where the observation function $h$ is in a Hilbert space on the system state $\mathbf{u}^{n}$ of the discrete system in Eq.~\eqref{eq:DTDS}. The Koopman operator is defined as an operator  $\mathcal{K}$ that evolves the observation function in that Hilbert space, i.e., $\mathcal{K}h := h \circ \mathcal{G}$, where $\circ$ denotes function composition. In practice, it can often be characterized (or approximated) by a finite-dimensional dynamical system in the form $\mathbf{v}(\mathbf{u}^{n+1}) = \mathbf{A}\mathbf{v}(\mathbf{u}^{n})$, where $\mathbf{v}$ is a vector function on the space of $\mathbf{u}^{n}$, $\mathbf{v}(\mathbf{u}^{n}) \in \mathbb{R}^{d_{\mathbf{v}}}$ is a finite-dimensional vector, and $\mathbf{A} \in \mathbb{R}^{d_{\mathbf{v}} \times d_{\mathbf{v}}}$ is a matrix. Obtaining such a finite-dimensional linear dynamical system for a general nonlinear system in Eq.~\eqref{eq:DTDS} involves the identification of the subspace spanned by eigenvectors of the Koopman operator $\mathcal{K}$, which has been studied in~\citep{lusch2018deep} via a deep learning perspective. More details on Koopman theory can be found in \ref{app:Koop}.

In this work, instead of directly applying the Koopman theory to the nonlinear dynamical system in Eq.~\eqref{eq:DTDS}, we exploit the linear embedding enabled by the Koopman theory on the partially observed system in Eq.~\eqref{eq:DTDS_u1u2} to obtain a conditional linear system, i.e., the governing equations become conditionally linear with respect to a latent representation $\mathbf{v}$ of the unobserved states $\mathbf{u}_2$, while the nonlinearity of observed states $\mathbf{u}_1$ is preserved. More details of generalized application of Koopman theory can be found in \ref{app:GenKoop}. The resulting modeled system is written as follows:
\begin{equation}
\begin{aligned}
\label{eq:NeuralCGNS}
\mathbf{u}_1^{n+1} &= \mathbf{F}_1\big(\mathbf{u}_1^{n}\big) + \mathbf{G}_1\big(\mathbf{u}_1^{n}\big)\mathbf{v}^{n} + \boldsymbol{\sigma}_1\boldsymbol{\epsilon}_1^{n},\\
\mathbf{v}^{n+1} &= \mathbf{F}_2\big(\mathbf{u}_1^{n}\big) + \mathbf{G}_2\big(\mathbf{u}_1^{n}\big)\mathbf{v}^{n} + \boldsymbol{\sigma}_2\boldsymbol{\epsilon}_2^{n},
\end{aligned}
\end{equation}
where $\mathbf{v} \in \mathbb{R}^{d_{\mathbf{v}}}$ denotes a latent representation of unobserved states $\mathbf{u}_2$ in Eq.~\eqref{eq:DTDS_u1u2} and is referred to as latent states. In this modeled system, $\mathbf{F}_1$, $\mathbf{G}_1$, $\mathbf{F}_2$, and $\mathbf{G}_2$ are four (potentially nonlinear) maps of observed states $\mathbf{u}_1$. The $\boldsymbol{\epsilon}_1 \in \mathbb{R}^{d_1}$ and $\boldsymbol{\epsilon}_2 \in \mathbb{R}^{d_{\mathbf{v}}}$ are independent Gaussian white noises with $\boldsymbol{\sigma}_1 \in \mathbb{R}^{d_1 \times d_1}$ and $\boldsymbol{\sigma}_2 \in \mathbb{R}^{d_{\mathbf{v}} \times d_{\mathbf{v}}}$ as the noise coefficients. The two noise terms represent model errors that quantify the approximation quality of the surrogate model to the true underlying system. Despite the use of Gaussian white noises, it is worth highlighting that the nonlinearity in the model can still capture highly non-Gaussian features, thereby compensating for model errors arising from various approximations. In this work, the four nonlinear maps $\mathbf{F}_1$, $\mathbf{G}_1$, $\mathbf{F}_2$, and $\mathbf{G}_2$, together with the nonlinear maps between unobserved states $\mathbf{u}_2$ and the latent states $\mathbf{v}$, are parameterized by neural networks and calibrated based on data of the original dynamical system in Eq.~\eqref{eq:DTDS_u1u2}. The modeled system in Eq.~\eqref{eq:NeuralCGNS} belongs to a class of dynamical systems called conditional Gaussian nonlinear system (CGNS), i.e., the conditional distribution $p(\mathbf{v}^{n} | \{\mathbf{u}_1^{i}\}_{i=0}^{n})$ is Gaussian, which facilitates efficient analytical formulae of DA (see \ref{app:CGNS}). Another advantage of the surrogate model is the latent space embedding, which exploits the possible intrinsic lower-dimensional structure of the unobserved states $\mathbf{u}_2$ and thus can further facilitate the computational efficiency of state forecast and DA. A schematic overview of the surrogate model is shown in Fig.~\ref{fig:SchematicDiagram}(a).

It is worth noting that $\mathbf{v}$ is constructed as a latent representation of $\mathbf{u}_2$, and its interaction with $\mathbf{u}_1$ is facilitated by the conditional Gaussian structure in Eq.~\eqref{eq:NeuralCGNS}, to allow for the use of analytical DA formulae. The modeled system in Eq.~\eqref{eq:NeuralCGNS} is related to the CGKN framework proposed in~\citep{chen2025cgkn} and can be viewed as a discrete version of the original CGKN framework. Unlike the continuous form of CGKN framework studied in~\citep{chen2025cgkn}, the discrete form in Eq.~\eqref{eq:NeuralCGNS} avoids the numerical integration along time with a small step size and thus enables the exploration of its performance on PDE-governed problems with high-dimensional (or even infinite-dimensional) system states. The choice of temporal step size involves a trade-off: a smaller step enables high temporal resolution capture of fine-scale dynamics, whereas a larger step is more suitable for modeling coarse-grained or long-term behaviors. The remainder of this section introduces the architecture of CGKN in its discrete-in-time form, the application of CGKN to state forecast and DA, the training of CGKN based on offline data, and the uncertainty quantification for the results of CGKN.

\subsection{CGKN Architecture}
\label{ssec:CGKN_Arch}

\begin{figure}[H]
    \centering
    \includegraphics[width=\linewidth]{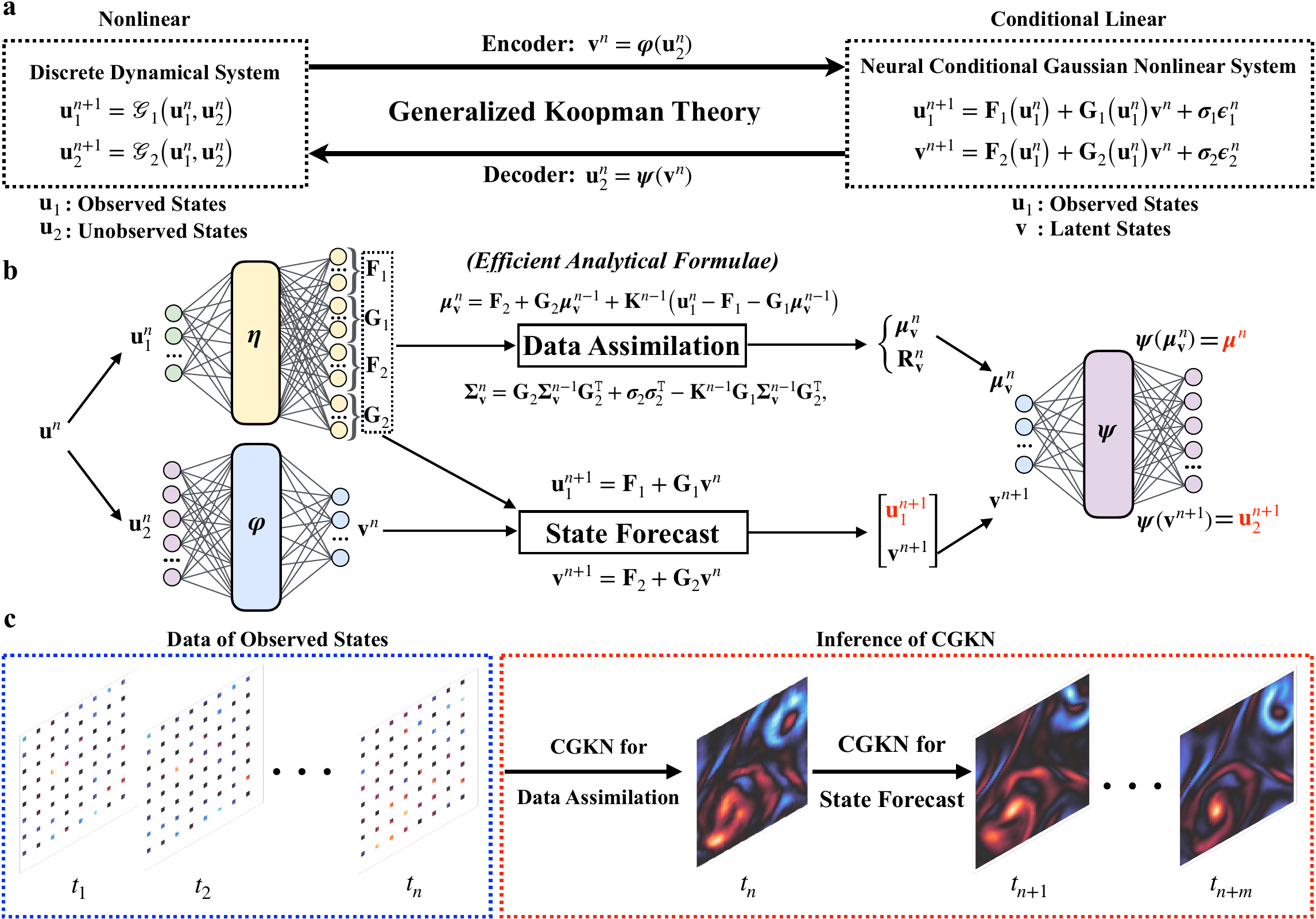}
    \caption{Schematic diagram of the architecture and application of the conditional Gaussian Koopman network (CGKN). \textbf{a}, Overview of the transformation from discrete dynamical systems to surrogate models following conditional Gaussian structure via generalized application of Koopman theory. \textbf{b}, The architecture and workflow of CGKN. The CGKN, consisting of an encoder $\boldsymbol{\varphi}$, a decoder $\boldsymbol{\psi}$ and sub-networks $\boldsymbol{\eta}$, is developed to learn the surrogate model for performing efficient DA via analytical formulae and state forecast for original dynamical systems.  \textbf{c}, An application of CGKN to Navier-Stokes equations for data assimilation and state forecast.}
    \label{fig:SchematicDiagram}
\end{figure}

As illustrated in Fig.~\ref{fig:SchematicDiagram}, the CGKN is developed to construct a surrogate model in Eq.~\eqref{eq:NeuralCGNS} to approximate the discrete dynamical system in Eq.~\eqref{eq:DTDS_u1u2}. The CGKN comprises an encoder $\boldsymbol{\varphi}$, a decoder $\boldsymbol{\psi}$, and sub-networks $\boldsymbol{\eta}$ to construct the nonlinear maps $\mathbf{F}_1$, $\mathbf{G}_1$, $\mathbf{F}_2$, and $\mathbf{G}_2$. The encoder $\boldsymbol{\varphi}$ is a nonlinear mapping for transforming the unobserved states $\mathbf{u}_2$ to the latent states $\mathbf{v}$, i.e., $\mathbf{v} = \boldsymbol{\varphi}(\mathbf{u}_2)$, while the decoder $\boldsymbol{\psi}$ is its inverse, i.e., $\mathbf{u}_2 = \boldsymbol{\psi}(\mathbf{v})$.

\subsection{CGKN for Forecast and DA of Discrete Dynamical Systems}
\label{ssec:CGKN_App}

\subsubsection{CGKN for State Forecast}
\label{sssec:CGKN_SF}

Assuming that the values of current states are known, the modeled system in Eq.~\eqref{eq:NeuralCGNS} can be used to forecast the future states of spatiotemporal dynamical systems. Here we denote the data of the state variables at time $t_n$ as $\mathbf{u}^{n\star} = \{\mathbf{u}_1^{n\star}, \mathbf{u}_2^{n\star}\}$, where the superscript $\star$ indicates the data from the true system in Eq.~\eqref{eq:DTDS_u1u2}. Firstly, the unobserved states $\mathbf{u}_2^{n\star}$ are transformed into latent states $\mathbf{v}^{n\star}$ by the encoder $\boldsymbol{\varphi}$, formalizing the current state of the modeled system $\mathbf{u}^{n\star}_{\textrm{CG}}$:
\begin{equation}
\label{eq:u_CG_Initial}
\begin{aligned}
\mathbf{u}_{\textrm{CG}}^{n\star}  :=
    \begin{bmatrix}
        \mathbf{u}^{n\star}_1 \\
        \mathbf{v}^{n\star}
    \end{bmatrix} =
        \begin{bmatrix}
        \mathbf{u}^{n\star}_1 \\
        \boldsymbol{\varphi}\big(\mathbf{u}^{n\star}_2\big)
    \end{bmatrix}.
\end{aligned}
\end{equation}

The forecast states $\mathbf{u}_{\textrm{CG}}$ after time $t_n$ can then be obtained from Eq.~$\eqref{eq:NeuralCGNS}$ by solving an initial value problem. It is worth noting that the predictive latent states $\mathbf{v}$ in the forecast states $\mathbf{u}_{\textrm{CG}}$ need to be transformed back to the original unobserved states $\mathbf{u}_2$, which is achieved via the decoder $\boldsymbol{\psi}$.

In practice, the data of the latent states $\mathbf{v}^{n\star}$ is often not available due to the lack of unobserved states data $\mathbf{u}_2^{n\star}$, which accounts for a key motivation of performing DA. Generally speaking, DA is a commonly used technique to leverage the data of observed states $\mathbf{u}_1$ to improve the estimation of $\mathbf{v}$, by obtaining the posterior distribution $p(\mathbf{v}^{n}|\{\mathbf{u}^{i}_1\}_{i=0}^n)$, and its maximum posterior point can serve as an estimation of the unknown latent states $\mathbf{v}^{n\star}$. If the Gaussian assumption is adopted in the DA methods, e.g., Kalman filters, the maximum posterior point becomes the mean value $\boldsymbol{\mu}_{\mathbf{v}}$ of the posterior distribution. 

Instead of merely employing DA methods (e.g., ensemble Kalman filter) as a separate task to improve the state estimation of nonlinear models, an important contribution of this work is to exploit the conditional linear structure of the modeled system in Eq.~\eqref{eq:NeuralCGNS} and to adopt the analytical formulae of DA that facilitate an efficient evaluation of the DA performance, which allows the incorporation of the DA performance into the training of a scientific machine learning model and thus enables a unified framework for scientific machine learning and data assimilation. More details of the analytical formulae of DA and the incorporation of DA performance into the model training are discussed in Sections~\ref{sssec:CGKN_DA} and~\ref{ssec:CGKN_Training}, respectively.

\subsubsection{CGKN for DA}
\label{sssec:CGKN_DA}

A key feature of the CGKN is that the conditional distribution $p(\mathbf{v}^{n} | \{\mathbf{u}_1^{i}\}_{i=0}^{n})$ is Gaussian, whose mean $\boldsymbol{\mu}_{\mathbf{v}}$ and covariance $\boldsymbol{\Sigma}_{\mathbf{v}}$ can be solved by the analytical formulae:
\begin{equation}
\label{eq:CGFilter}
    \begin{aligned}
    \boldsymbol{\mu}_{\mathbf{v}}^{n+1} &= \mathbf{F}_2 + \mathbf{G}_2\boldsymbol{\mu}_{\mathbf{v}}^{n} + \mathbf{K}^{n} \big(\mathbf{u}_1^{n+1} - \mathbf{F}_1 - \mathbf{G}_1\boldsymbol{\mu}_{\mathbf{v}}^{n}\big),\\
    \mathbf{\boldsymbol{\Sigma}}_{\mathbf{v}}^{n+1} &= \mathbf{G}_2\mathbf{\boldsymbol{\Sigma}}_{\mathbf{v}}^{n}\mathbf{G}_2^\mathtt{T} + \boldsymbol{\sigma}_2\boldsymbol{\sigma}_2^\mathtt{T} - \mathbf{K}^{n}\mathbf{G}_1\mathbf{\boldsymbol{\Sigma}}^{n}_{\mathbf{v}}\mathbf{G}_2^\mathtt{T},
    \end{aligned}
\end{equation}
where $\mathbf{K}^{n} = \mathbf{G}_2 \mathbf{\boldsymbol{\Sigma}}_{\mathbf{v}}^{n}\mathbf{G}_1^\mathtt{T}\big(\boldsymbol{\sigma}_1\boldsymbol{\sigma}_1^\mathtt{T} + \mathbf{G}_1\mathbf{\boldsymbol{\Sigma}}_{\mathbf{v}}^{n}\mathbf{G}_1^\mathtt{T}\big)^{-1}$ is an analog to the Kalman gain. The mean $\boldsymbol{\mu}_{\mathbf{v}}$ of the latent states $\mathbf{v}$ can be transformed to the posterior mean $\boldsymbol{\mu}$ of unobserved states $\mathbf{u}_2$ by the decoder $\boldsymbol{\psi}$, i.e., $\boldsymbol{\mu} = \boldsymbol{\psi}(\boldsymbol{\mu}_{\mathbf{v}})$. It is worth noting that such a transformation of the estimated mean $\boldsymbol{\mu}_{\mathbf{v}}$ generally does not hold, but the DA loss term used to train the CGKN model can ensure the validity of the posterior mean $\boldsymbol{\mu}$. More discussions about the DA loss term can be found in Section~\ref{ssec:CGKN_Training}. On the other hand, the posterior covariance $\boldsymbol{\Sigma}$ of the unobserved states $\mathbf{u}_2$ can be estimated through residual analysis, and additional information on the uncertainty quantification can be found in Section~\ref{sssec:UQ_DA}. It should be noted that the initial condition of Eq.~\eqref{eq:CGFilter} is typically unknown. Therefore, a warm-up period is required to mitigate initialization discrepancy.

The analytical formulae in Eq.~\eqref{eq:CGFilter}, together with the lower dimensionality of latent state ($d_{\mathbf{v}} \ll d_2$) when an intrinsic lower-dimensional representation of unobserved states is possible, account for two key features of CGKN that enable efficient DA compared with ensemble DA methods (e.g., ensemble Kalman filter). The computational complexity of CGKN for DA is $\mathcal{O}(d_\mathbf{v}^3)$, where $d_\mathbf{v}$ is the dimension of the latent states, while that of ensemble DA methods is $\mathcal{O}(Jd_2^2)$, where $J$ is the ensemble size and $d_2$ is the dimension of unobserved states. To mitigate the sampling errors in ensemble DA methods, a large ensemble size may be needed to well characterize the covariance structure, and thus we often have $Jd_2^2 \gg d_\mathbf{v}^3$, making the proposed CGKN framework a more efficient choice for DA than ensemble-based methods. It is worth noting that a small ensemble size $J$ is often employed in practice even for large-scale problems, e.g., numerical weather prediction (NWP) commonly uses an ensemble size on the order of $10^2$, despite the dimension of system state being several orders of magnitude larger. The use of a small ensemble size for large-scale problems, together with various techniques such as localization~\citep{tong2023localized,liu2023dropout} or damping small correlations~\citep{vishny2024high}, is still an active research area for ensemble DA methods.

\subsection{Learning CGKN from Data}
\label{ssec:CGKN_Training}

For the training of CGKN model, we assume that offline data $\{\mathbf{u}^{n\star}\}_{n=0}^N$ of both observed states $\mathbf{u}_1$ and unobserved ones $\mathbf{u}_2$ are available from the true system in Eq.~\eqref{eq:DTDS}. The encoder parameters $\boldsymbol{\theta}_{\boldsymbol{\varphi}}$, decoder parameters $\boldsymbol{\theta}_{\boldsymbol{\psi}}$ and sub-networks parameters  $\boldsymbol{\theta}_{\boldsymbol{\eta}}$ of the CGKN model illustrated in Fig.~\ref{fig:SchematicDiagram}(b) are optimized by minimizing a total loss function:
\begin{equation}
\label{eq:opt_problem}
\min_{\boldsymbol{\theta}_{\boldsymbol{\varphi}},\boldsymbol{\theta}_{\boldsymbol{\psi}}, \boldsymbol{\theta}_{\boldsymbol{\eta}}}  L(\boldsymbol{\theta}_{\boldsymbol{\varphi}},\boldsymbol{\theta}_{\boldsymbol{\psi}}, \boldsymbol{\theta}_{\boldsymbol{\eta}}),
\end{equation}
which includes autoencoder loss $L_{\mathrm{AE}}$, forecast loss of original states $L_{\mathbf{u}}$, forecast loss of latent states $L_{\mathbf{v}}$, and DA loss $L_{\mathrm{DA}}$:
\begin{equation}
\label{eq:Loss_Target}
L(\boldsymbol{\theta}_{\boldsymbol{\varphi}},\boldsymbol{\theta}_{\boldsymbol{\psi}}, \boldsymbol{\theta}_{\boldsymbol{\eta}}) := \lambda_\textrm{AE} L_{\textrm{AE}}(\boldsymbol{\theta}_{\boldsymbol{\varphi}},\boldsymbol{\theta}_{\boldsymbol{\psi}}) + \lambda_{\mathbf{u}} L_{\mathbf{u}}(\boldsymbol{\theta}_{\boldsymbol{\varphi}},\boldsymbol{\theta}_{\boldsymbol{\psi}}, \boldsymbol{\theta}_{\boldsymbol{\eta}}) + \lambda_{\mathbf{v}} L_{\mathbf{v}}(\boldsymbol{\theta}_{\boldsymbol{\varphi}}, \boldsymbol{\theta}_{\boldsymbol{\eta}}) + \lambda_{\textrm{DA}}L_{\textrm{DA}}(\boldsymbol{\theta}_{\boldsymbol{\psi}}, \boldsymbol{\theta}_{\boldsymbol{\eta}}),
\end{equation}
where
\begin{align}
L_{\textrm{AE}} &:= \mathbb{E}_{\mathbf{u}_2^\star}\|\mathbf{u}_2^\star - \boldsymbol{\psi}\big(\boldsymbol{\varphi}(\mathbf{u}_2^\star)\big)\|^2,\label{eq:Loss_AE}\\
L_{\mathbf{u}} &:= \mathbb{E}_{\mathbf{u}^{0\star}}\frac{1}{N_s}\sum_{n=1}^{N_s} \|\mathbf{u}^{n\star} -  \mathbf{u}^n\|^2,\label{eq:Loss_u}\\
L_{\mathbf{v}} &:= \mathbb{E}_{\mathbf{u}^{0\star}}\frac{1}{N_s}\sum_{n=1}^{N_s} \|\mathbf{v}^{n\star} -  \mathbf{v}^{n}\|^2,\label{eq:Loss_v}\\
L_{\textrm{DA}} &:= \mathbb{E}_{\mathbf{u}^{0\star}}\frac{1}{N_l-N_b}\sum_{\substack{n=N_b+1}}^{N_l} \|\mathbf{u}^{n\star}_2 - \boldsymbol{\mu}^{n}\|^2,\label{eq:Loss_DA}
\end{align}
with $\lambda_{\textrm{AE}}$, $\lambda_{\mathbf{u}}$, $\lambda_{\mathbf{v}}$, $\lambda_{\textrm{DA}}$ being the weights for each loss term. $\|\cdot\|$ denotes the standard vector $\ell^2$-norm. The $N_s$, $N_l$, and $N_b$ are three hyper-parameters in the training settings which stand for forecast steps, DA steps, and DA warm-up steps, respectively. The $\mathbf{u}^n$ and $\mathbf{v}^n$ are the $n$-th step state predictions which is described in Section~\ref{sssec:CGKN_SF}, and $\boldsymbol{\mu}^n$ is $n$-th step DA posterior mean which is detailed in Section~\ref{sssec:CGKN_DA}. In this work, the weights is set as $\lambda_{\textrm{AE}} = 1/d_{2}$, $\lambda_{\mathbf{u}} = 1/d$, $\lambda_{\mathbf{v}} = 1/d_{\mathbf{v}}$, and $\lambda_{\textrm{DA}} = 1/d_{2}$, which scale each loss function by a factor of the inverse of the respective vector's dimension. This results in loss functions expressed as the mean squared error (MSE) between the true and approximated data. In practice, the weights can also be treated as hyper-parameters or manually adjusted based on the domain knowledge.

The autoencoder loss $L_{\textrm{AE}}$ in Eq.~\eqref{eq:Loss_AE} is used to minimize the reconstruction discrepancy between the unobserved states $\mathbf{u}^{n\star}_2$ (as the input of autoencoder) and $\boldsymbol{\psi}\big(\boldsymbol{\varphi}(\mathbf{u}^{n\star}_2)\big)$ (as the output of autoencoder). The loss $L_{\textrm{AE}}$ promotes the inverse relationship between the encoder and decoder, which helps identify latent states $\mathbf{v} = \boldsymbol{\varphi}(\mathbf{u}_2)$ which can be regarded as embeddings of unobserved states. It is worth noting that the latent space to approximately embed unobserved states is usually not unique, and the joint usage of $L_{\textrm{AE}}$ and other loss terms facilitate the discovery of a proper latent space that is favored by the goals of having good performance in the forecast and DA results of the CGKN model. More specifically, the forecast loss of original states $L_{\mathbf{u}}$ and forecast loss of latent states in $L_{\mathbf{v}}$ in Eqs.~\eqref{eq:Loss_u} and~\eqref{eq:Loss_v} are incorporated to ensure that CGKN makes accurate state forecast, while the DA loss $L_{\textrm{DA}}$ in Eq.~\eqref{eq:Loss_DA}, made affordable by the efficient DA formulae in Eq.~\eqref{eq:CGFilter}, is critical to ensure the accuracy of DA results from CGKN. It should be noted that the magnitude of $\mathbf{v}$ and $\mathbf{G}_1$ in Eq.~\eqref{eq:NeuralCGNS} could be scaled without impacting system states $\mathbf{u}_1$, and a corresponding scaling of the autoencoder can leave system states $\mathbf{u}_2$ unaffected. In this work, such a scale invariance of $\mathbf{v}$ would not lead to any issue, as $\mathbf{v}$ is only a latent representation, and the system states $\mathbf{u}_1$ and $\mathbf{u}_2$ are of actual interest. In practice, the magnitude of $\mathbf{v}$ is also impacted by the choice of $\boldsymbol{\sigma}_2$ in Eq.~\eqref{eq:NeuralCGNS}.

The forecast loss of original states $L_{\mathbf{u}}$ and forecast loss of latent states $L_{\mathbf{v}}$ defined in Eq.~\eqref{eq:Loss_u} and Eq.~\eqref{eq:Loss_v} is incorporated to enable CGKN to make accurate state forecast for dynamical systems. From the true states $\{\mathbf{u}^{n\star}\}_{n=0}^{N_s}$ with $\mathbf{u}^{n\star} = \{\mathbf{u}_1^{n\star}, \mathbf{u}_2^{n\star}\}$, the corresponding latent states $\{\mathbf{v}^{n\star}\}_{n=0}^{N_s}$ can be obtained from unobserved states via encoder: $\mathbf{v}^{n\star} = \boldsymbol{\varphi}\big(\mathbf{u}_2^{n\star}\big)$. Based on the initial states $\{\mathbf{u}_1^{0\star}, \mathbf{v}^{0\star}\}$, the predictive states $\{\mathbf{u}_1^{n}, \mathbf{v}^{n}\}$ can be obtained by recursively evaluating the surrogate model as illustrated in Section~\ref{sssec:CGKN_SF}. The predictive unobserved states are transformed from the predictive latent states via decoder: $\mathbf{u}_2^n = \boldsymbol{\psi}\big(\mathbf{v}^n\big)$. $L_{\mathbf{u}}$ measures the discrepancy between true states $\{\mathbf{u}^{n\star}\}_{n=0}^{N_s}$ and predictive states $\{\mathbf{u}^n\}_{n=0}^{N_s}$. $L_{\mathbf{v}}$ measures the discrepancy between true latent states $\{\mathbf{v}^{n\star}\}_{n=0}^{N_s}$ and predictive latent sates $\{\mathbf{v}^n\}_{n=0}^{N_s}$. The $N_s$ is the forecast horizon, which can be set as a hyper-parameter in the training setting.

The DA loss $L_{\textrm{DA}}$ in Eq.~\eqref{eq:Loss_DA}, which is made affordable by the efficient DA formulae in Eq.~\eqref{eq:CGFilter}, is critical to ensure the accuracy of DA results from CGKN. Based on the data of observed states $\{\mathbf{u}_1^{n\star}\}_{n=0}^{N_l}$, the posterior mean of the latent states $\{\boldsymbol{\mu}_{\mathbf{v}}^n\}_{n=0}^{N_l}$ are calculated by applying the analytical DA formulae in Eq.~\eqref{eq:CGFilter}. The posterior mean of unobserved states $\{\boldsymbol{\mu}^n\}_{n=0}^{N_l}$ is obtained by the decoder: $\boldsymbol{\mu}^n = \boldsymbol{\psi}\big(\boldsymbol{\mu}_{\mathbf{v}}^n\big)$. The DA loss $L_{\textrm{DA}}$ quantifies the discrepancy between the true unobserved states $\{\mathbf{u}_2^{n\star}\}_{n=0}^{N_l}$ and DA posterior mean $\{\boldsymbol{\mu}^n\}_{n=0}^{N_l}$, excluding the first $N_b$ steps. $N_l$ represents the number of DA steps, and $N_b$ denotes the DA warm-up steps, both of which are hyper-parameters in the training setting. The role of DA loss has been investigated in \citep{chen2024cgnsde}.

It should be noted that the analytical DA formulae in Eq.~\eqref{eq:CGFilter} require the value of  $\boldsymbol{\sigma}_1$ and $\boldsymbol{\sigma}_\mathbf{2}$ in the surrogate model in Eq.~\eqref{eq:NeuralCGNS}. Therefore, prior to incorporating the DA loss into the target loss function to train CGKN, the $\boldsymbol{\sigma}_1$, which is assumed to be a diagonal matrix, is estimated by root mean squared error (RMSE) between the true observed states and one-step predictive observed states via a trained CGKN without DA loss. On the other hand, the $\boldsymbol{\sigma}_2$, which is the noise coefficient of latent states, is on the bias term in the analytical DA formulae in Eq.~\eqref{eq:CGFilter}. In practice, $\boldsymbol{\sigma}_2$ can be set either manually or as a trainable parameter jointly learned with all the trainable maps in the CGKN model. We have studied different choices of $\boldsymbol{\sigma}_2$, and the results of the learned CGKN models are robust based on those numerical tests. The detailed results are omitted here for simplicity. A brief algorithm about the procedures of learning CGKN from data is summarized in Algorithm~\ref{alg:CGKN}. More details about the uncertainty quantification are introduced in the following section.

\subsection{Uncertainty Quantification}
\label{ssec:UQ}

This section discusses the uncertainty quantification (UQ) for state forecast and DA results of CGKN. The task of UQ for state forecast focuses on the estimation of $\boldsymbol{\sigma}_1$, which is the noise coefficient of the observed states $\mathbf{u}_1$ and $\boldsymbol{\sigma}_2$, which is the noise coefficient of the latent states $\mathbf{v}$. The task of UQ for DA is to estimate the posterior covariance of the conditional distribution $p(\mathbf{u}^{n}_2 | \{\mathbf{u}_1^{i}\}_{i=0}^{n})$, which is generally non-Gaussian. Both UQ tasks are post-processes of pre-trained CGKN models: the task of UQ for DA is based on the CGKN model trained with the DA loss term, while the task of UQ for state forecast relies on the CGKN model trained without the DA loss term. More details are introduced in the following sections.

\subsubsection{Uncertainty Quantification for State Forecast}
\label{sssec:UQ_SF}

To perform the task of UQ for state forecast, a CGKN model with the total loss in the form of Eq.~\eqref{eq:Loss_Target} while excluding the DA loss term needs to be trained, i.e., the weight of DA loss $L_{\textrm{DA}}$ in the target loss function $L$ in Eq.~\eqref{eq:Loss_Target} is set as $\lambda_{\textrm{DA}}=0$. Once having the pre-trained CGKN model without DA loss, the noise coefficient $\boldsymbol{\sigma}_1$ of observed states and noise coefficient $\boldsymbol{\sigma}_2$ of latent states are assumed to be diagonal matrices. The diagonal elements are estimated by the root mean squared error (RMSE) between the true states and one-step predictive states of the pre-trained CGKN model:
\begin{equation}
\label{eq:sigma_estimation}
    \begin{aligned}
        \mathrm{diag}(\boldsymbol{\sigma}_1) &= \sqrt{ \frac{1}{N}\sum_{n=1}^{N} \bigl(\mathbf{u}_1^{n\star} -  \mathbf{u}_1^{n}\bigr) \odot \bigl(\mathbf{u}_1^{n\star} -  \mathbf{u}_1^{n}\bigr)},  \\
        \mathrm{diag}(\boldsymbol{\sigma}_2) &= \sqrt{ \frac{1}{N}\sum_{n=1}^{N} \bigl(\mathbf{v}^{n\star} -  \mathbf{v}^{n}\bigr) \odot \bigl(\mathbf{v}^{n\star} -  \mathbf{v}^{n}\bigr)},
    \end{aligned}
\end{equation}
where the notation $\odot$ is element-wise multiplication. Additionally, the estimated $\boldsymbol{\sigma}_1$ and $\boldsymbol{\sigma}_2$ in Eq.~\eqref{eq:sigma_estimation} can be utilized to calculate the DA loss through analytical DA formulae in Eq.~\eqref{eq:CGFilter}, which facilitates unifying the training of the CGKN model and the tuning of its DA performance based on the total loss in Eq.~\eqref{eq:Loss_Target} with all loss terms included. It is worth noting that an accurate a priori estimation of those noise coefficients may not be necessary to ensure a good tuning of DA performance, since there are other trainable maps in the analytical DA formulae in Eq.~\eqref{eq:CGFilter}. In practice, the noise coefficients can also be empirically set or jointly learned with those trainable maps of the CGKN model.

\subsubsection{Uncertainty Quantification for Data Assimilation}
\label{sssec:UQ_DA}

As introduced in Section~\ref{ssec:CGKN_Training}, the DA loss term measures the discrepancy between the true unobserved states $\mathbf{u}_2$ and the posterior mean $\boldsymbol{\mu}$, which promotes the performance of DA in terms of mean estimation. However, the posterior covariance of the unobserved states $\mathbf{u}_2$, which quantifies the uncertainty of the associated posterior mean, is still unaddressed. Since true covariance data for the unobserved states does not exist for many real-world applications, especially for those only with experimental data available, it is not feasible to apply a supervised learning framework to estimate the posterior covariance in a similar fashion as being done for the posterior mean.

Therefore, residual analysis is exploited as a post-processing method for a trained CGKN, to efficiently quantify the uncertainties associated with the posterior mean of the unobserved states $\mathbf{u}_2$. Residual analysis is commonly used to evaluate the goodness of a statistical model with respect to observed data. With a trained CGKN model, the DA can be performed for the training data via analytical formulae in Eq.~\eqref{eq:CGFilter} to obtain the posterior mean $\boldsymbol{\mu}_{\mathbf{v}}$. With the trained decoder $\boldsymbol{\psi}$, we can further obtain posterior mean $\boldsymbol{\mu}$ of unobserved states $\mathbf{u}_2$. The residual $\mathbf{r}$ is defined as the absolute difference between true data $\mathbf{u}_2^{\star}$ and its DA posterior mean $\boldsymbol{\mu}$:
\[
\mathbf{r} := |\mathbf{u}_2^{\star} - \boldsymbol{\mu}|, 
\]
which indicates the desired uncertainties associated with the posterior mean $\boldsymbol{\mu}$ and is assumed to be a function of observed states $\mathbf{u}_1$. This function is then approximated by an auxiliary neural network trained via a standard regression task from the pairwise dataset $\{\big(\mathbf{u}^{n\star}_1, \mathbf{r}^{n}\big)\}_{n=N_b+1}^N$, with the first $N_b$ steps as the warm-up period of DA.

With the trained auxiliary neural network for UQ, the residual associated with the posterior mean can be estimated, using the corresponding observed states as input and evaluating the output of the auxiliary neural network. This residual, derived from the trained auxiliary neural network, can be regarded as the estimated standard deviation of the associated posterior mean, i.e., the standard deviation of diagonal elements of the posterior covariance, based on maximum likelihood estimation under the Gaussian assumption.

\begin{algorithm}[H]
\caption{Learning CGKN from data}
\label{alg:CGKN}
\begin{algorithmic}
        \State \textbf{Input}: $\{\mathbf{u}^{n\star}\}_{n=0}^N$  \Comment{Training data}
        \State  $\boldsymbol{\theta}_{\boldsymbol{\varphi}}^{(1)}, \boldsymbol{\theta}_{\boldsymbol{\psi}}^{(1)}, \boldsymbol{\theta}_{\boldsymbol{\eta}}^{(1)} = \argmin \big\{\lambda_{\textrm{AE}}L_{\textrm{AE}} + \lambda_{\mathbf{u}}L_{\mathbf{u}} + \lambda_{\mathbf{v}}L_{\mathbf{v}}\big\}$  \Comment{Train CGKN without DA loss}
        \State $\mathrm{diag}(\boldsymbol{\sigma}) = \{\mathrm{diag}(\boldsymbol{\sigma}_1), \mathrm{diag}(\boldsymbol{\sigma}_2)\}  = \textrm{RMSE}(\mathbf{u}^\star_\textrm{CG}, \mathbf{u}_\textrm{CG})$ \Comment{UQ for state forecast}
        \State  $\boldsymbol{\theta}_{\boldsymbol{\varphi}}^{(2)}, \boldsymbol{\theta}_{\boldsymbol{\psi}}^{(2)}, \boldsymbol{\theta}_{\boldsymbol{\eta}}^{(2)} = \argmin \big\{\lambda_{\textrm{AE}}L_{\textrm{AE}} + \lambda_{\mathbf{u}}L_{\mathbf{u}} + \lambda_{\mathbf{v}}L_{\mathbf{v}} + \lambda_{\textrm{DA}}L_{\textrm{DA}} \big\}$  \Comment{Train CGKN with DA loss}
        \State $\boldsymbol{\theta}^*_{\textrm{UQ}} = \argmin \textrm{MSE}\big(\mathbf{r}, \textrm{NN}(\mathbf{u}^\star_1;\boldsymbol{\theta}_{\textrm{UQ}})\big)$ \Comment{UQ for DA}
        \State \textbf{Output}: $\boldsymbol{\theta}_{\boldsymbol{\varphi}}^{(2)}, \boldsymbol{\theta}_{\boldsymbol{\psi}}^{(2)}, \boldsymbol{\theta}_{\boldsymbol{\eta}}^{(2)}, \boldsymbol{\sigma}, \boldsymbol{\theta}^*_{\textrm{UQ}}$ \Comment{Trained parameters}
\end{algorithmic}
\end{algorithm}

\section{Numerical Experiments}
\label{sec:NumericalExperiments}

The effectiveness and efficiency of CGKN in state forecast and DA for spatiotemporal dynamical systems are demonstrated through numerical experiments on several PDE-governed canonical examples and benchmarks against other models. The examples include the viscous Burgers' equation, the Kuramoto-Sivashinsky equation, and the Navier-Stokes equations. The methods that have been tested for each system are summarized as follows:
\begin{enumerate}[(i)]
    \item CGKN. It is the deep learning framework proposed in this work, which formalizes the surrogate model in Eq.~\eqref{eq:NeuralCGNS} to perform state forecast and DA for spatiotemporal dynamical systems.

    \item Ensemble Kalman filter (EnKF) \citep{evensen1994sequential, Anderson2001EAKF}. EnKF is a classical method for applying DA to nonlinear dynamical systems. The localization strategy and empirical tunings \citep{anderson_monte_1999, gaspari1999construction} are often used to enhance the EnKF's performance. The comparison of DA results with the CGKN model highlights the accuracy and efficiency of the CGKN model.
    
    \item Direct spatial interpolation. Based on the values of observed states across the spatial domain, interpolation can be utilized to estimate the missing values of unobserved states, recovering the entire fields. It can be viewed as a naive approach to estimate the full-field information based on spatially sparse observations, and it serves as a baseline reference solution for comparing the DA results from CGKN and EnKF.

    \item Deep neural network (DNN). The model of DNN with fully-connected layers is calibrated to approximate the unknown map $\mathcal{G}$ on the right-hand side in Eq.~\eqref{eq:DTDS}, and it serves as one of the baseline benchmarks for the performance of state forecast.

    \item Convolutional neural network (CNN) \citep{lecun1995convolutional, lecun1989handwritten}. The unknown map $\mathcal{G}$ on the right-hand side in Eq.~\eqref{eq:DTDS} is approximated using a CNN model, which serves as one of the baseline benchmarks for the performance of state forecast.

    \item Fourier neural operator (FNO). FNO has been demonstrated in~\citep{li2021fourier} with superior performance than classical machine learning methods (e.g., fully-connected networks and CNN) for PDE-governed problems whose system states are spatiotemporal fields. In this work, the unknown map $\mathcal{G}$ on the right-hand side in Eq.~\eqref{eq:DTDS} is approximated by an FNO to predict future states. Although it is expected that FNO can outperform all the previous neural-network-based models (including CGKN) for the performance of state forecast, the key advantage of CGKN is its efficient DA.

\end{enumerate}

To compare the performance of the methods above, the mean squared error (MSE) is used to evaluate the performance of state forecast and DA:
\begin{equation}
    \label{eq:MSE}
    \textrm{MSE} := \frac{1}{MN} \sum_{m=1}^{M} \sum_{n=1}^N \big( \mathbf{x}_m^{n\star} - \mathbf{x}_m^{n} \big)^2,
\end{equation}
where $\mathbf{x}^{n\star}_m$ is the true value of the $m$-th variable at the $n$-th time step, for $m=1,...,M$ and $n=1,...,N$, and $\mathbf{x}^n_m$ is its corresponding approximated value. The DA error is defined as the MSE between the true unobserved states and their posterior mean from DA, while the forecast error is defined as the MSE between true states and the predictive ones. Both DA error and forecast error are calculated based on test data. Additionally, the forecast error is calculated based on one-step prediction. while the DA error is calculated over the whole time range of the test dataset. The test results of all methods across all examples are summarized in Table~\ref{tab:Test_Errors}. For a fair comparison, a comparable number of parameters is used in all those deep-learning-based models.

\begin{table}[H]
\caption{Test results of all methods in each numerical example. The errors are mean squared errors (MSE) between true values and approximated values. For a fair comparison, a comparable number of parameters is used in different deep learning models.}
\label{tab:Test_Errors}
\begin{adjustbox}{max width=1.\textwidth,center}
\begin{tabular}{|c|c|c|c|c|c|c|}
\hline
\multirow{2}{*}{\diagbox{Methods}{Examples}} & \multicolumn{2}{c|}{Viscous Burgers Equation} & \multicolumn{2}{c|}{Kuramoto–Sivashinsky Equation} & \multicolumn{2}{c|}{Navier–Stokes Equations} \\
\cline{2-7}
& Forecast Error & DA Error & Forecast Error & DA Error  & Forecast Error & DA Error \\
\hline
CGKN & 7.5683e-04 & 7.5037e-04 & 1.1042e-02 & 2.4927e-02 & 1.9754e+01 & 6.0940e+01 
 \\
\hline
EnKF & --- & 5.8125e-04 & ---  & 2.4882e-02 & --- & 6.9010e+01   \\
\hline
Interpolation & --- & 1.3514e-02 & --- & 4.3097e-01  & --- & 1.2844e+02       \\
\hline
DNN & 6.4816e-03 & ---  & 4.7332e-02 & ---  & 1.0936e+02  & ---  \\
\hline
CNN & 2.3727e-03 & ---  & 2.6111e-02 & ---  & 3.0600e+01  & ---  \\
\hline
FNO & 3.9715e-04 & --- & 5.4859e-03  & ---  & 1.7129e+01 & ---   \\
\hline
\end{tabular}
\end{adjustbox}
\end{table}

\subsection{Viscous Burgers' Equation: 1-D PDE with Shock Behavior}
\label{ssec:VBE}

The 1-D Burgers' equation is a fundamental nonlinear PDE from fluid mechanics, with many applications in various fields such as gas dynamics, traffic flow, and acoustics. It is widely used for studying complex fluid behaviors, particularly notable for modeling shock waves and Burgulence phenomena. The form of the viscous Burgers' equation is:
\begin{equation}
\begin{aligned}
\dfrac{\partial u}{\partial t} = -u\dfrac{\partial u}{\partial x} + \nu \dfrac{\partial^2 u}{\partial x^2},
\end{aligned}
\label{eq:VBE}
\end{equation}
where $x \in [0, L_x]$ with periodic boundary conditions, $t \in [0, L_t]$, $\nu$ is the viscosity coefficient, and $u(x,0)$ is a given initial condition. The simulation settings are $L_x = 1, L_t = 2, \nu=10^{-3}, \Delta x = 1/1024,  \Delta t = 10^{-3}$. Under these conditions, the viscous Burgers' equation displays shock behavior due to the formation of steep gradients in the evolution. We perform 1000 simulations as train data and another 100 simulations as test data. The initial conditions $u(x,0)$ of these simulations are randomly sampled from Gaussian process $\mathcal{N}(0, 625(-\Delta +25I)^{-2})$ with periodic boundary conditions where $\Delta$ is a Laplace operator and $I$ is an identity operator. The resolution of both the train and the test dataset is sub-sampled to $\Delta t=0.1$ and $\Delta x=1/64$. Among the 64 states, 4 states which are uniformly distributed across the spatial domain are set as observed, and the rest 60 states are set as unobserved. More specifically, the indices of observed states are 1, 17, 33, 49. The CGKN has been used to learn a surrogate model for i) estimating the 60 unobserved states from the trajectory of the sparse 4 observations via efficient DA and ii) forecasting the future state of the system given any initial state.

To approximate the state transition map of the viscous Burgers' equation from data using the surrogate model in Eq.~\eqref{eq:NeuralCGNS}, the CGKN including an encoder $\boldsymbol{\varphi}$, a decoder $\boldsymbol{\psi}$, and sub-networks $\boldsymbol{\eta}$ that output $\mathbf{F}_1$, $\mathbf{G}_1$, $\mathbf{F}_2$, $\mathbf{G}_2$ has been constructed. With the dimension of latent states $d_\mathbf{v}$ set to $10$, the encoder, which takes the unobserved states $\mathbf{u}_2 \in \mathbb{R}^{60}$ as input and output latent states $\mathbf{v} \in \mathbb{R}^{10}$, become $\boldsymbol{\varphi}: \mathbb{R}^{60} \mapsto \mathbb{R}^{10}$, and its inverse, the decoder, become $\boldsymbol{\psi}: \mathbb{R}^{10} \mapsto \mathbb{R}^{60}$. Consequently, the observed states $\mathbf{u}_1 \in \mathbb{R}^{4}$ and latent states $\mathbf{v}$ formalize the new system states of the surrogate model in Eq.~\eqref{eq:NeuralCGNS}. Furthermore, the nonlinear components $\mathbf{F}_1 \in \mathbb{R}^{4}$, $\mathbf{G}_1 \in \mathbb{R}^{4\times 10}$, $\mathbf{F}_2 \in \mathbb{R}^{10} $, $\mathbf{G}_2 \in \mathbb{R}^{10\times 10}$ in the surrogate model are output by the sub-networks $\boldsymbol{\eta}:\mathbb{R}^{4} \mapsto \mathbb{R}^{128}$, which take observed states $\mathbf{u}_1$ as input. For each component of CGKN, the number of parameters $\boldsymbol{\theta}_{\boldsymbol{\varphi}}$, $\boldsymbol{\theta}_{\boldsymbol{\psi}}$, and $\boldsymbol{\theta}_{\boldsymbol{\eta}}$ are 17066, 17116, 34330, respectively. To learn the CGKN from data, the training parameters are set as follows: state forecast horizon $t_{N_s} = 2$ time units (20 steps), the DA horizon $t_{N_l} = 1000$ time units (1000 steps), and a warm-up period $t_{N_b} = 0.8$ time units (8 steps). The weights in the target loss function in Eq.~\eqref{eq:Loss_Target} are set as $\lambda_{\textrm{AE}} = 1/d_{2}$, $\lambda_{\mathbf{u}}= 1/d$, $\lambda_{\mathbf{v}} = 1/d_{\mathbf{v}}$, and $\lambda_{\textrm{DA}} = 1/d_2$.

The test results of state forecast and DA from all methods for the viscous Burgers' equation are summarized in Table~\ref{tab:Test_Errors} under the column viscous Burgers' equation. The EnKF is applied on the governing equations of viscous Burgers' equation in Eq.~\eqref{eq:VBE} with the prior knowledge of the initial distribution of Gaussian process. The ensemble size of EnKF is set as $J=100$ and the initial ensembles are randomly sampled from the same Gaussian process that generates the initial conditions for training data. To ensure stability and accuracy in the numerical simulation of the governing equations, a temporal resolution of $\Delta t=0.01$ has been employed and a spatial resolution of $\Delta x=1/256$ has been linearly interpolated from the data. Based on the empirical tunings from training data, neither inflation nor localization strategies are needed when applying the EnKF for the test data in this example. 

The CGKN can achieve the same level of accuracy in DA as the EnKF, which applies to the true governing equation, and both methods significantly outperform simple interpolation. Regarding time cost, the CGKN requires 0.02 seconds to perform DA for a single simulation (20 steps), compared to 12 seconds for the EnKF. Although these time costs are based on this numerical example and the computing device used in this work, the CGKN is expected to be more computationally efficient in data assimilation than EnKF for a much wider range of problems and computing devices. It should be noted that the numerical simulation for each ensemble member of the viscous Burgers' equation is sequential. While the time cost of EnKF could be reduced through parallel computing, it would require more memory resources. The time usage confirms the speed advantages of CGKN, which are attributed to the efficient DA formulae, the reduced dimensionality of the latent states, and fast forecast through the approximation of the state transition map. Additionally, as a data-driven method, the DA from CGKN does not require any empirical tunings and any prior knowledge such as the initial distribution of Gaussian process and the governing equation. 

As for the performance of state forecast, the DNN is the worst mainly because the fully-connected layer is less effective at leveraging spatial information. The CGKN outperforms the CNN in state forecast and approaches the results achieved by the FNO, which demonstrates the effectiveness of embedding-based learning in CGKN. It is worth highlighting that all general nonlinear neural-network-based models, including DNN, CNN, and FNO, are not able to perform efficient and accurate DA due to their black-box nature and lack of the constraint of DA loss in the training stage. The numerical results demonstrate that the CGKN is effective in both state forecast and efficient DA for the viscous Burger's equation.

\begin{figure}[H]
    \centering
    \includegraphics[width=0.9\linewidth]{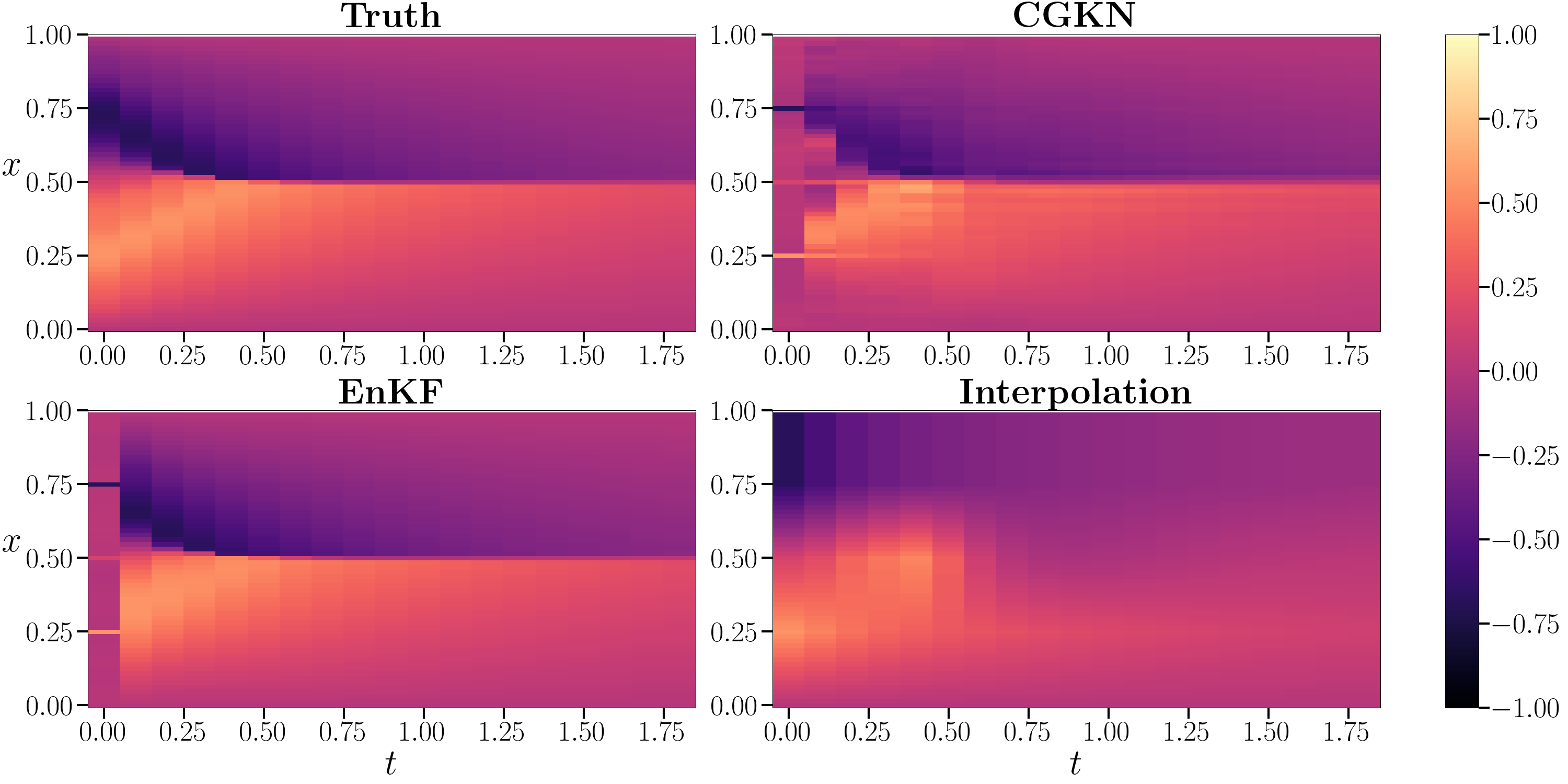}
    \caption{Spatiotemporal evolution of the DA results for the viscous Burgers' equation. The spatiotemporal plots of true simulation, DA posterior mean from CGKN, DA posterior mean from EnKF, and interpolation result are shown in each sub-figure. EnKF is applied to the true governing equation with the knowledge of the prior initial distribution.}
    \label{fig:VBE_DA_Pcolor}
\end{figure}

Figure~\ref{fig:VBE_DA_Pcolor} shows the spatiotemporal plot of the true simulation of the viscous Burgers' equation, DA posterior means from CGKN and EnKF, and the interpolation result. In the results from the true system, the shock behavior is characterized by the horizontal line in the middle part that indicates a dramatic change in the values of the system state within a small spatial distance. Starting with an initial guess, both the flows of CGKN and EnKF can gradually adjust to match the true flow by assimilating observations from the four observed states. In contrast, the pattern of interpolation results deviates significantly from the truth. Additionally, the position of shock behavior can be well-captured by both CGKN and EnKF, while it is not explicit from the interpolation.

\begin{figure}[H]
    \centering
    \includegraphics[width=\linewidth]{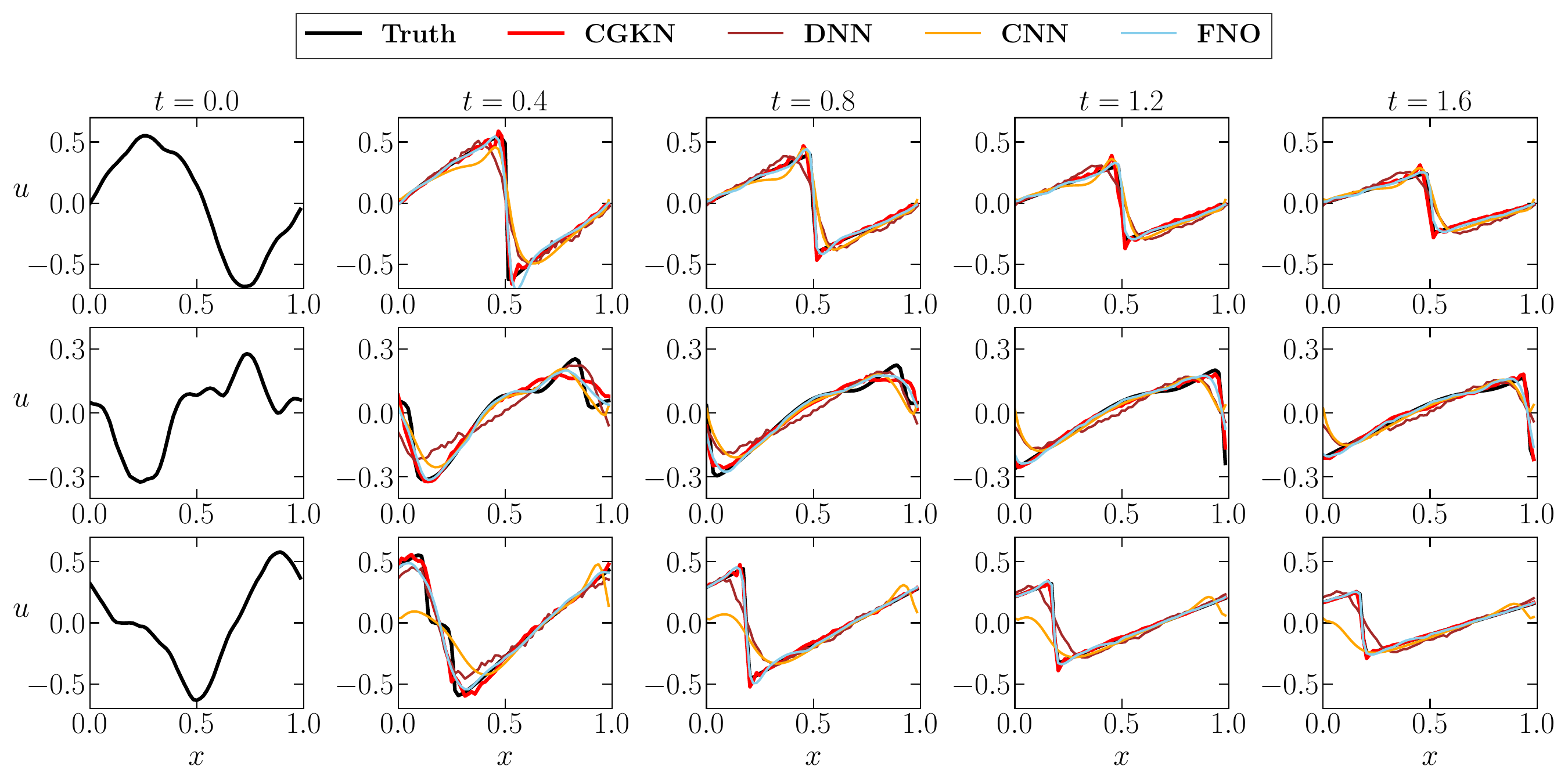}
    \caption{Results of state forecast for the viscous Burger's equation. Starting from three different initial conditions in test data, the evolution of true spatial functions is compared with predictive spatial functions from various models, including CGKN, DNN, CNN, and FNO.}
    \label{fig:VBE_SF_SpatialProfile}
\end{figure}

The results of the state forecast from various neural-network-based models and the true simulation are shown in Fig.~\ref{fig:VBE_SF_SpatialProfile}, using three different initial conditions from the test dataset. To evaluate the performance of state forecast of those models, we assume the initial condition is known and then numerically simulate the modeled systems for a certain amount of time. The predictions from DNN and CNN display noticeable deviations from the true spatial functions, while the CGKN and FNO can match the truth well and capture the shock behavior. This figure illustrates the comparable state forecast performance of CGKN and FNO, highlighting CGKN's capability in state forecast. Compared to all the general nonlinear neural-network-based models, the key advantage of CGKN is its capability of efficient DA, which achieves similar performance as applying EnKF to the true governing equations, while demanding much less computational resources in both tuning and performing DA.

\begin{figure}[H]
    \centering
    \includegraphics[width=0.8\linewidth]{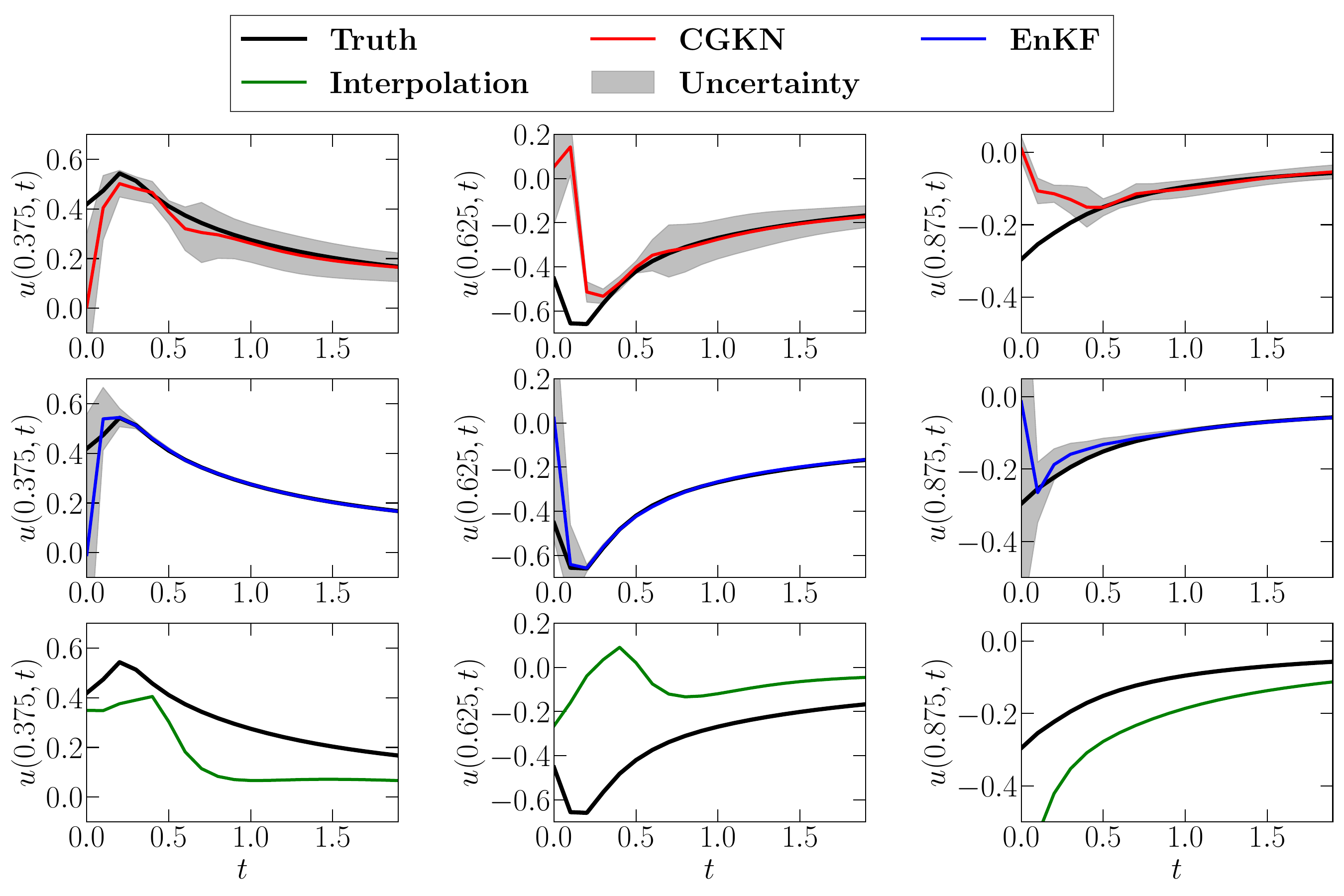}
    \caption{Time series of the DA results for the viscous Burgers' equation. True signals, DA posterior means from CGKN and EnKF together with uncertainty areas, and interpolation results for three unobserved states are shown in the figure. The uncertainties of the two standard deviations are indicated by the gray-colored regions associated with the posterior mean.}
    \label{fig:VBE_DA_TimeSeries}
\end{figure}

In many real-world applications, the initial condition is often unknown or only partially known, which accounts for a key motivation of performing DA. Figure~\ref{fig:VBE_DA_TimeSeries} displays the DA results of unobserved states at the locations $x=0.375$, $0.624$, and $0.875$ in the spatial domain. The posterior means with uncertainties characterized by two standard deviations from CGKN and EnKF are compared with the true signals in the first two rows, while the interpolation result is compared with the true signals in the last row. The uncertainty from CGKN is derived from the method of residual analysis, while that from EnKF is obtained from the empirical standard deviation estimated from ensemble samples. Starting from the initial guess, the posterior means from both CGKN and EnKF can match the true signal after about 0.2 time units (2 steps). Though both the uncertainty areas can cover the true signals, the uncertainty area from CGKN is wider than that from EnKF. For the interpolation results, there is a significant deviation from the true signals throughout the entire time range of flow evolution.

\begin{figure}[H]
    \centering
    \includegraphics[width=\linewidth]{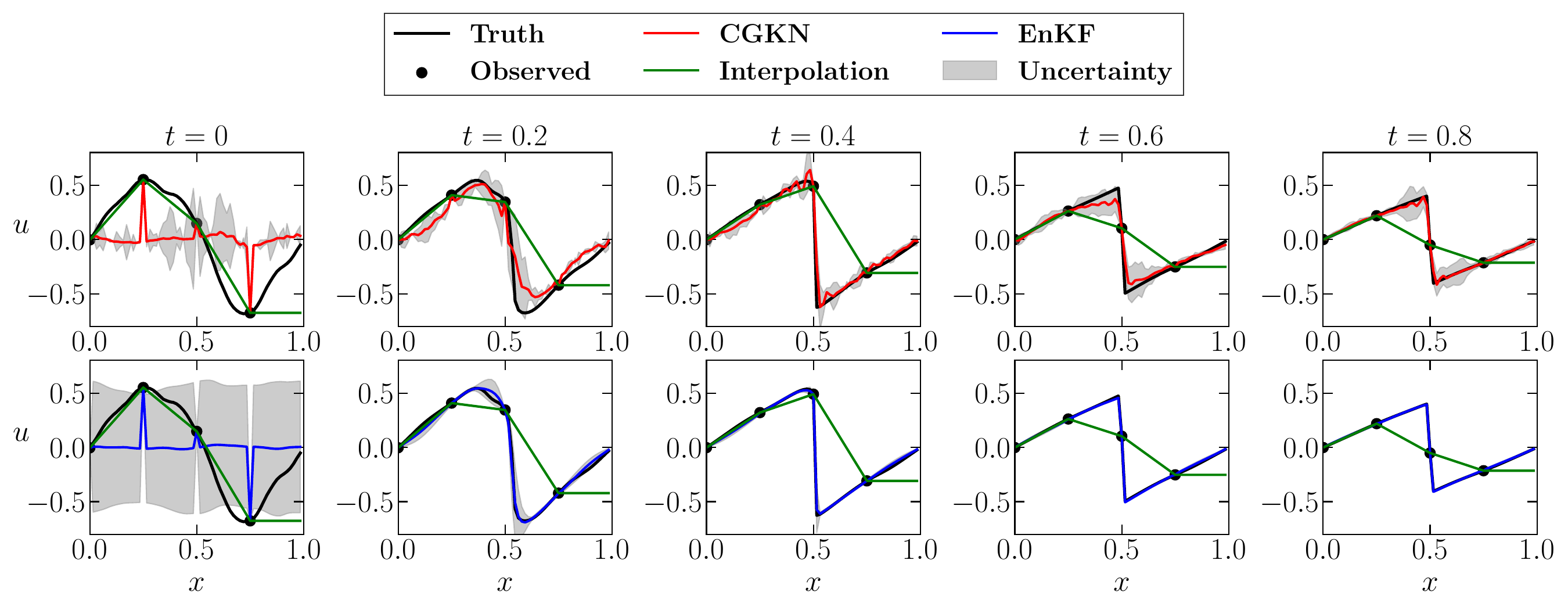}
    \caption{Spatial profiles of the DA results for the viscous Burgers' equation. Starting from a random guess, the DA results from CGKN are compared with the true spatial functions in the first row, while those from EnKF are displayed in the second row. The interpolation result is shown in both rows as a reference.}
    \label{fig:VBE_DA_SpatialProfile}
\end{figure}

Figure~\ref{fig:VBE_DA_SpatialProfile} compares the entire solution profile evolving over time with the DA results obtained from EnKF and CGKN, as well as the interpolation result. It can be seen that a quick adaptation of the DA results to the true signal is achieved via applying EnKF to the true system. At the time $t=0.2$, the posterior mean from EnKF can almost match the true signal, while there are still some deviations from that of CGKN. The CGKN results gradually align with the true signal in subsequent time steps. It is expected that applying EnKF to the true system can lead to almost perfect state estimation results in subsequent time steps. On the other hand, it is worth mentioning that the posterior mean from CGKN may not always precisely match the true signal at some points, mainly due to the trade-off between fitting a data-driven model and ensuring its generalization capability, especially in regions close to the steep gradient in space. Nevertheless, the uncertainty area still encompasses the true signal for most of those regions, thereby enhancing the reliability of the prediction results.

\subsection{Kuramoto-Sivashinsky Equation: 1-D Chaotic PDE}
\label{ssec:KSE}

The Kuramoto–Sivashinsky (K-S) equation is a fourth-order nonlinear PDE that was originally developed to model diffusive-thermal instabilities in a laminar flame front and appears in various areas of physics and engineering such as reaction-diffusion systems. It has been extensively studied for its wave-like dynamics and chaotic behavior, serving as a prototypical example of a system that exhibits spatiotemporal chaos. The form of the Kuramoto–Sivashinsky equation is:
\begin{equation}
\label{eq:KSE}
\begin{aligned}
    \frac{\partial u}{\partial t} = -u\frac{\partial u}{\partial x} - \frac{\partial^2 u}{\partial x^2} - \frac{\partial^4 u}{\partial x^4},
\end{aligned}
\end{equation}
where $x \in [0, L_x]$ with periodic boundary conditions and $t \in [0, L_t]$. The simulation settings are $L_x=22, L_t=5000, \Delta x=22/2048, \Delta t=0.025$ and the initial condition is $u(x,0) = 0.1\times\cos(x/16)\times(1+2\sin(x/16))$. Under the settings, the K-S equation displays a strong chaotic behavior, making it a good testbed for the proposed method. In the simulation of 5000 time units, the first $80\%$ of the simulation (4000 time units) is used as train data and the remaining $20\%$ (1000 time units) is used as test data. Both train and test data are sub-sampled to resolution $\Delta t=1$ and $\Delta x=22/128$. To simulate the real scenarios and demonstrate the robustness of the proposed method, the observational noise following $\mathcal{N}(0, 0.2^2)$ has been independently and identically added to the true states, including both train and test datasets. The observed states are located at 8 points uniformly distributed across the 1-D spatial domain, whereas the remaining 120 points are unobserved. More specifically, the indices of observed states are 1, 17, 33, 49, 65, 81, 97, 113. The application of CGKN for this 1-D chaotic spatiotemporal system aims to i) estimate the $128$ states across the spatial domain from a trajectory data of $8$ observed states and ii) forecast the $128$ states in the future given any initial $128$ states.

With an encoder $\boldsymbol{\varphi}$, a decoder $\boldsymbol{\psi}$, and sub-networks $\boldsymbol{\eta}$ that output $\mathbf{F}_1$, $\mathbf{G}_1$, $\mathbf{F}_2$, $\mathbf{G}_2$, the surrogate model in Eq.~\eqref{eq:NeuralCGNS} is constructed to approximate the state transition map of the K-S equation based on the data resolution. The dimension of latent state $d_\mathbf{v}$ is selected as 12 in this example. Consequently, the encoder, which take unobserved states $\mathbf{u}_2 \in \mathbb{R}^{120}$ as input and output latent states $\mathbf{v} \in \mathbb{R}^{12}$, becomes $\boldsymbol{\varphi}: \mathbb{R}^{120} \mapsto \mathbb{R}^{12}$ and decoder becomes $\boldsymbol{\psi}: \mathbb{R}^{12} \mapsto \mathbb{R}^{120}$. The observed states $\mathbf{u}_1 \in \mathbb{R}^{8}$ and latent state $\mathbf{v}$ formalize the system state of the surrogate model in Eq.~\eqref{eq:NeuralCGNS}. The sub-networks $\boldsymbol{\eta}:\mathbb{R}^{8} \mapsto \mathbb{R}^{260}$ takes observed states $\mathbf{u}_1$ as input and outputs the nonlinear components $\mathbf{F}_1 \in \mathbb{R}^{8}$, $\mathbf{G}_1 \in \mathbb{R}^{8\times 12}$, $\mathbf{F}_2 \in \mathbb{R}^{12} $, $\mathbf{G}_2 \in \mathbb{R}^{12\times 12}$ in the surrogate model. The number of parameters $\boldsymbol{\theta}_{\boldsymbol{\varphi}}$, $\boldsymbol{\theta}_{\boldsymbol{\psi}}$, and $\boldsymbol{\theta}_{\boldsymbol{\eta}}$ of each component in CGKN are 18422, 18530, and 37956, respectively. The training settings for the CGKN include: state forecast horizon $t_{N_s} = \Delta t = 1$ time unit which is the temporal resolution of data (i.e., one-step prediction) and the DA horizon $t_{N_l} = 1000$ time units (1000 steps) with the warm-up period set as $t_{N_b} = 5$ time units (5 steps). The weights in the target loss function are set as $\lambda_{\textrm{AE}} = 1/d_2$, $\lambda_{\mathbf{u}}= 1/d$, $\lambda_{\mathbf{v}} = 1/d_{\mathbf{v}}$, and $\lambda_{\textrm{DA}} = 1/d_2$.

The test results for state forecast and DA from different methods are summarized in Table~\ref{tab:Test_Errors} under the column Kuramoto-Sivashinsky equation. The DA error is based on 1000-time-unit (1000 steps) horizon, and the forecast error is calculated by one-step prediction. The interpolation serves as a basic reference model for the DA results from CGKN and EnKF. The EnKF is applied to the true governing equation of K-S equation in Eq.~\eqref{eq:KSE}. The ensemble size of EnKF is set as $J=100$ with the initial ensemble randomly sampled from the training data, in contrast to the initial distribution of Gaussian noise used by DA of CGKN. The constant multiplicative covariance inflation \citep{anderson_monte_1999} and the localization strategy \citep{gaspari1999construction} are used to mitigate sampling errors. Based on the empirical tunings for training data, the inflation and localization parameters are set as 1.05 and 16. The numerical simulation of the governing equations for the forecast step in EnKF employs a smaller temporal resolution of $\Delta t=0.025$ than the data temporal resolution to ensure numerical stability and accuracy. 

For the DA performance, the CGKN is comparable to applying EnKF on the true governing equations, and both methods are significantly superior to the interpolation. Based on the same computing device, the CGKN completes 1000 steps of DA for test data in about 0.8 seconds, while EnKF requires about 100 seconds to accomplish the same task. Although the EnKF can be accelerated via parallel computing for the numerical simulation of ensemble members, it will cost more memory resources. The efficiency of DA from CGKN compared to EnKF stems from the analytical DA formulae, dimension reduction of the latent state, and fast forecast of the approximated state transition map. Therefore, even though the DA performance of CGKN is comparable to that of EnKF, it can significantly reduce computational costs or require fewer computing resources. 

For the state forecast, the FNO outperforms all other models and the DNN performs the worst. The performance of CGKN is better than that of CNN and is only slightly less effective than the FNO, a neural-operator-based architecture specifically designed for learning mappings between infinite-dimensional spaces. It should be emphasized that, due to their black-box nature and the absence of DA loss in the training stage, general nonlinear neural network models—including DNNs, CNNs, and FNOs—are not specifically designed for efficient and accurate DA tasks. This highlights the advantage of CGKN.

\begin{figure}[H]
    \centering
    \includegraphics[width=\linewidth]{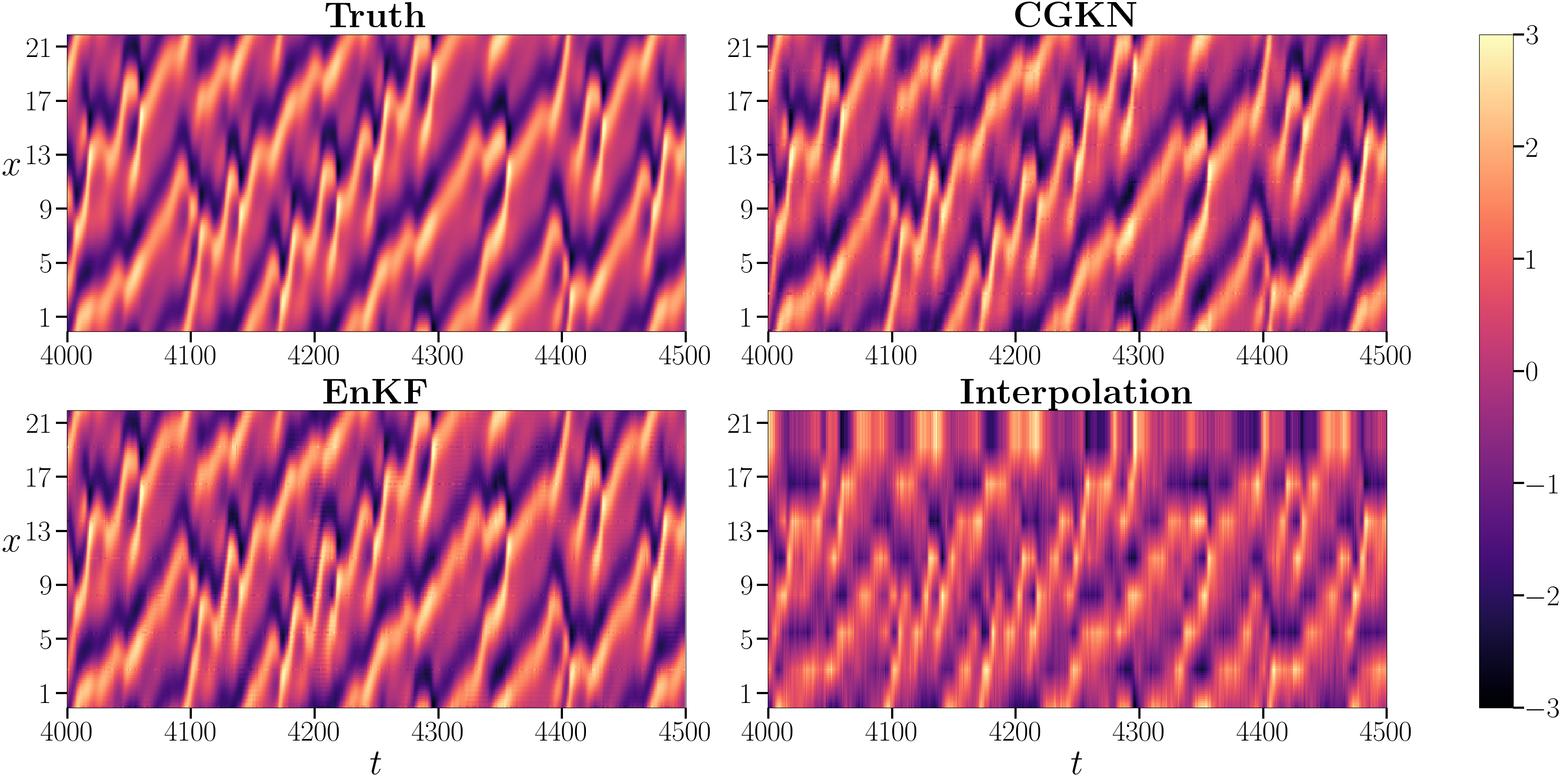}
    \caption{Spatiotemporal evolution of the DA results for the Kuramoto-Sivashinsky equation. The spatiotemporal plots of true simulation, DA posterior mean from CGKN, DA posterior mean from EnKF, and interpolation result are displayed in each sub-figure.}
    \label{fig:KSE_DA_Pcolor}
\end{figure}

Figure~\ref{fig:KSE_DA_Pcolor} displays the spatiotemporal plot of the true simulation of the K-S equation, posterior DA means from CGKN and EnKF, and interpolation result for 500 time units. The result from CGKN closely resembles that of EnKF, with both methods aligning well with the truth. Both DA methods can effectively capture chaotic patterns and wave-like behaviors. Conversely, the result from interpolation is noticeably different from the true values, although it does manage to qualitatively recover some wave-like patterns.

\begin{figure}[H]
    \centering
    \includegraphics[width=\linewidth]{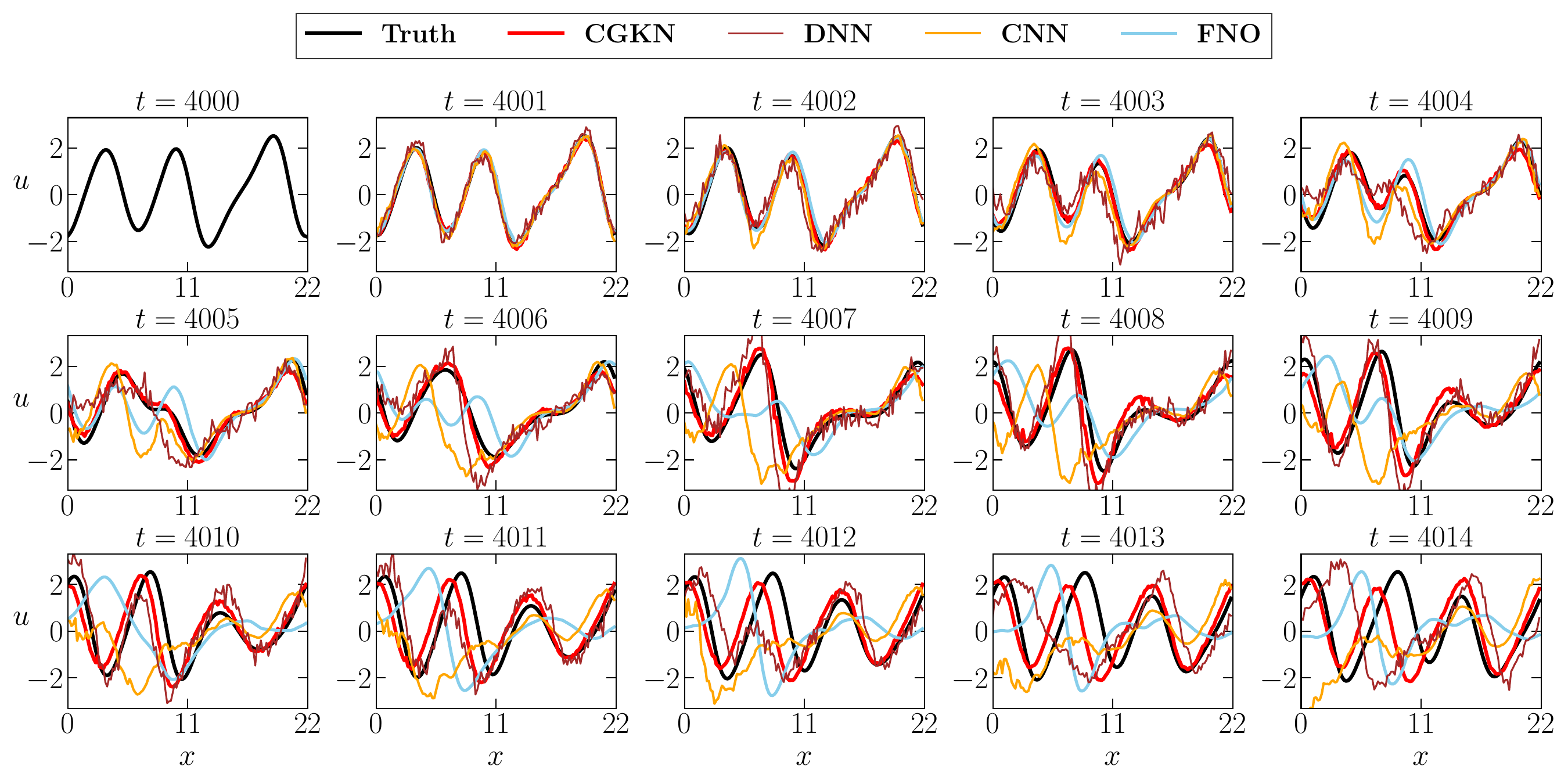}
    \caption{Results of state forecast for the Kuramoto-Sivashinsky equation. The evolution of true spatial functions is compared with predictive spatial functions from various models including CGKN, DNN, CNN, and FNO.}
    \label{fig:KSE_SF_SpatialProfile}
\end{figure}

Figure~\ref{fig:KSE_SF_SpatialProfile} illustrates the state forecast performance of various models by displaying the evolution of spatial functions of both the true and predictive results starting at 4000 time units, which is in the test dataset. For the one-step prediction, all models accurately match the true spatial function. However, for larger forecast steps, all models gradually diverge from the truth, with CGKN maintaining the closest approximation over the longest forecast horizon in this example. The comparable state forecast capability of CGKN to FNO stems from their similar modeling frameworks: CGKN models the state transition map in latent space, while FNO models that in Fourier space. The effectiveness of CGKN in state forecast primarily benefits from the informative embeddings it learns from data, a feature that is absent in the architectures of DNN and CNN. Similar to the previous numerical example, the key advantage of CGKN over those general nonlinear neural-network-based models is efficient DA, which achieves comparable performance as applying EnKF to the true governing equations but demands significantly fewer computational resources and does not require empirical tunings.

\begin{figure}[H]
    \centering
    \includegraphics[width=0.7\linewidth]{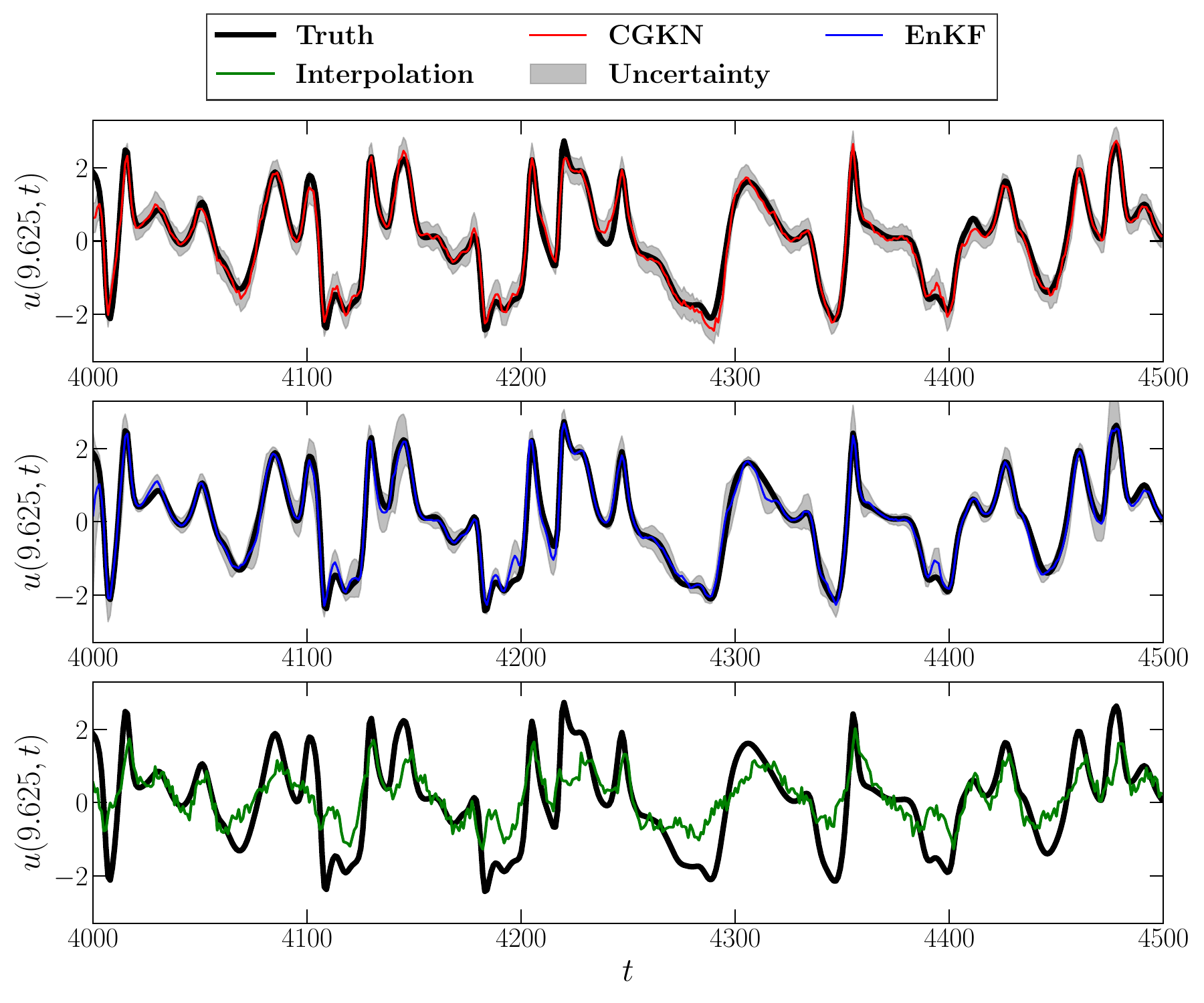}
    \caption{Time series of the DA results for the Kuramoto-Sivashinsky equation. True signal, DA posterior mean from CGKN and EnKF together with uncertainty area, and interpolation results for the unobserved state at spatial position 9.625 are presented. The uncertainties of the two standard deviations are indicated by the gray-colored regions associated with the posterior mean.}
    \label{fig:KSE_DA_TS}
\end{figure}

Figure~\ref{fig:KSE_DA_TS} presents the time series of the unobserved state at the location $x = 9.625$ in the spatial domain $[0, 22]$. The true signal of the unobserved state is compared with the DA posterior means obtained from CGKN and EnKF, as well as with the interpolation result. For both DA methods, the uncertainty area corresponding to two standard deviations is shown around the respective posterior means. The uncertainty from  CGKN is derived through residual analysis, while that from EnKF is the empirical standard deviation estimated from ensemble samples. The trajectories of the DA means from both CGKN and EnKF closely match the true signal, significantly outperforming the interpolation results. The uncertainty area provided by CGKN is similar to that of EnKF and covers most of the true signals. Additionally, the DA results from both CGKN and EnKF rapidly adjust to the system and accurately estimate the unobserved state from the initial guess, indicating that only a short warm-up time is required for this spatiotemporal chaotic system.

\begin{figure}[H]
    \centering
    \includegraphics[width=\linewidth]{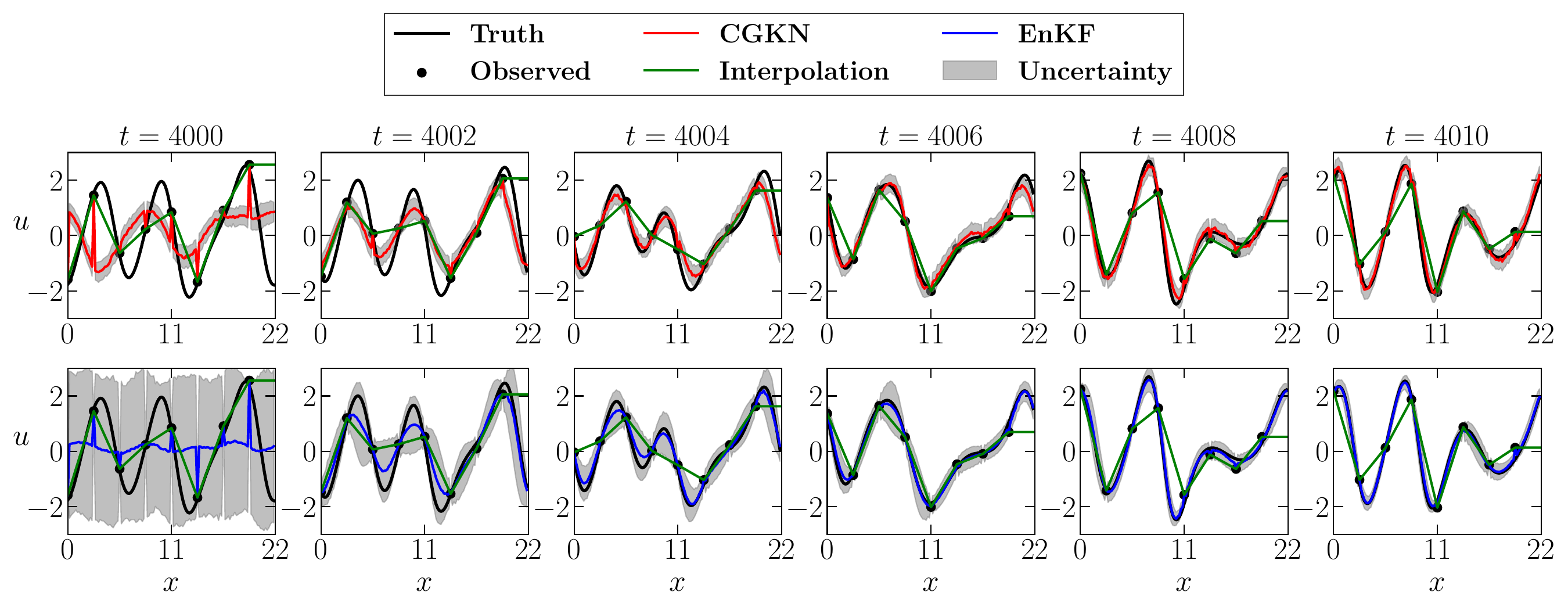}
    \caption{Spatial profiles of the DA results for the Kuramoto-Sivashinsky equation. Starting from a random guess, the DA results from CGKN are compared with the true spatial functions in the first row, while those from EnKF are presented in the second row. The interpolation results are provided as references in both rows.}
    \label{fig:KSE_DA_SpatialProfile}
\end{figure}

The initial phases of the DA processes from CGKN and EnKF are illustrated in Fig.~\ref{fig:KSE_DA_SpatialProfile}, which also includes the true spatial functions and interpolation results. The figure shows the spatial profiles corresponding to the spatiotemporal evolution presented in Fig.~\ref{fig:KSE_DA_Pcolor} for the first 10 time units. It is evident that both posterior means align well with the true spatial functions after approximately 6 time units, whereas the interpolation result continues to serve only as a baseline reference. Both Figs~\ref{fig:KSE_DA_TS} and \ref{fig:KSE_DA_SpatialProfile} confirm that CGKN achieves DA performance comparable to using EnKF with the true system.

\subsection{Navier-Stokes Equations: 2-D Turbulent PDE}
\label{ssec:NSE}

The Navier-Stokes (N-S) equations are a set of nonlinear PDEs that describe the motion of fluids. These equations are fundamental in the field of fluid dynamics and are used in a wide range of applications in both natural and engineering, including weather patterns, ocean currents, and airflow. The velocity form of the Navier–Stokes equations for incompressible fluid is:
\begin{equation}
\begin{gathered}
    \label{eq:NSE}
    \frac{\partial \mathbf{u}}{\partial t} = -\mathbf{u}\cdot \nabla \mathbf{u} + \nu \nabla^2 \mathbf{u} -\frac{1}{\rho}\nabla p + \mathbf{f}, \\
    \nabla \cdot \mathbf{u} = 0,
\end{gathered}
\end{equation}
where $\mathbf{u}(\mathbf{x}, t) \in \mathbb{R}^2$ is a 2-D velocity vector defined over the spatial domain $\mathbf{x} \in \Omega \subset \mathbb{R}^2$ and temporal domain $t \in [0, L_t]$. Here, $\nu$ is the kinematic viscosity, $\rho$ is the density, $p$ is the pressure field, and $\mathbf{f}$ is the forcing function. The state of interest in this example is the vorticity, which is the curl of velocity $\omega:=\nabla \times \mathbf{u}$. The simulations settings for the system are $\nu=10^{-4}$, $\rho=1$, $\mathbf{f} = [100\sin(8y), 0]^\mathtt{T}$, $\Omega=[0, 1]^2$, $L_t = 1000$, $\Delta t=10^{-3}$, and $\Delta x=\Delta y=1/256$. The initial condition is $\mathbf{u}(\mathbf{x}, 0) = \mathbf{0}$ and the boundary condition is periodic. The N-S equations with this setup result in a 2-D spatiotemporal turbulence. The stable fluids algorithm is used to solve the 2-D Navier-Stokes equations with the above settings to generate 1000-time-unit vorticity. The first 800 units are used as train data and the remaining 200 units are used as test data. To demonstrate the robustness of CGKN and simulate real-world applications, the measurement noise of $\mathcal{N}(0, 2.5^2)$ has been independently and identically added to both the train and test data. The resolution of both train and test data is sub-sampled to $\Delta t=0.01$, $\Delta x=1/64$ and $\Delta y=1/64$. In the $64 \times 64$ 2-D vorticity field, $8 \times 8$ observed points are set uniformly distributed across the spatial domain, while the other points are unobserved. The proposed CGKN for this 2-D turbulent vorticity field aims to i) estimate the $64 \times 64$ vorticity field from a trajectory of $8\times 8$ sparse observed data and ii) forecast the future $64 \times 64$ vorticity field given any initial $64 \times 64$ vorticity field.

With the CGKN that includes an encoder $\boldsymbol{\varphi}$, a decoder $\boldsymbol{\psi}$, and sub-networks $\boldsymbol{\eta}$ that output $\mathbf{F}_1$, $\mathbf{G}_1$, $\mathbf{F}_2$, $\mathbf{G}_2$, the surrogate model in Eq.~\eqref{eq:NeuralCGNS} is constructed to approximate the state transition map of the 2-D vorticity field. The hyper-parameter $d_\mathbf{v}$, which is the dimension of latent states, has been selected as $16\times 16 = 256$ in this example. To facilitate the generalized application of Koopman theory for this 2-D turbulence, the convolutional autoencoder has been employed with encoder $\boldsymbol{\varphi}: \mathbb{R}^{64\times 64} \mapsto \mathbb{R}^{16\times 16}$ and decoder $\boldsymbol{\psi}: \mathbb{R}^{16\times l6} \mapsto \mathbb{R}^{64\times 64}$. The states in the surrogate model are constituted by observed states $\mathbf{u}_1 \in \mathbb{R}^{64}$ and latent states $\mathbf{v} \in \mathbb{R}^{256}$. The sub-networks $\boldsymbol{\eta}:\mathbb{R}^{64} \mapsto \mathbb{R}^{82240}$ takes observed states $\mathbf{u}_1$ as input and outputs the nonlinear components $\mathbf{F}_1 \in \mathbb{R}^{64}$, $\mathbf{G}_1 \in \mathbb{R}^{64\times 256}$, $\mathbf{F}_2 \in \mathbb{R}^{256} $, $\mathbf{G}_2 \in \mathbb{R}^{256\times 256}$ in the surrogate model. The number of parameters $\boldsymbol{\theta}_{\boldsymbol{\varphi}}$, $\boldsymbol{\theta}_{\boldsymbol{\psi}}$, and $\boldsymbol{\theta}_{\boldsymbol{\eta}}$ in the CGKN are 187265,187265 and 82240, respectively. The training settings for the CGKN include: state forecast horizon $t_{N_s} = \Delta t = 0.01$ time units which is the temporal resolution of data (i.e., one-step prediction) and the DA horizon $t_{N_l} = 20$ time units (2000 steps) with the warm-up period set as $t_{N_b} = 5$ time units (500 steps). The weights in the target loss function are set as $\lambda_{\textrm{AE}} = 1/d_2$, $\lambda_{\mathbf{u}}= 1/d$, $\lambda_{\mathbf{v}} = 1/d_{\mathbf{v}}$, and $\lambda_{\textrm{DA}} = 1/d_2$.

The test results of state forecast and DA from different methods are summarized in Table~\ref{tab:Test_Errors} under the column Navier-Stokes equations. The DA error is based on a 200-time-unit (20000 steps) horizon and the forecast error is calculated by one-step prediction. The EnKF is applied on the true governing equations in Eq.~\eqref{eq:NSE} with the ensemble size set as $J=100$. Initial conditions of ensembles in EnKF are randomly sampled from the training data, while the initial distribution of CGKN is set as Gaussian white noise. The constant multiplicative covariance inflation and the localization strategy have been taken for EnKF to exploit spatial information and reduce the sampling errors. The parameters of inflation and localization are chosen as $1.1$ and $8$ based on the empirical tunings of training data. The numerical simulation of the governing equation for the forecast step in EnKF uses a temporal resolution of $\Delta t=0.001$, compared with $0.01$ of the data, to ensure numerical stability and accuracy. 

In this 2-D turbulence example, the DA performance of CGKN outperforms that of EnKF, with the interpolation result displaying an error twice as large as that observed in CGKN. Additionally, CGKN completes DA for 20000 steps of test data in about 80 seconds, whereas EnKF requires about 25000 seconds (about 7 hours) to perform the same task. Though the EnKF can be accelerated by parallel computing for numerical simulation of ensemble numbers, it will cost more memory resources, especially for this 2-D turbulence simulation. The efficient DA of CGKN stems from the availability of analytical DA formulae, the reduced dimensionality of latent state, and fast forecast through approximation of the state transition map. Overall, with substantial advantages in terms of computational cost, the DA result from CGKN is still comparable to that of EnKF.

For state forecast, CGKN achieves performance comparable to FNO and outperforms both CNN and DNN, with DNN delivering the worst results. This is primarily because the DNN fails to leverage spatial information, which is crucial for analyzing a 2-D field. The state forecast performance demonstrates the effectiveness of the embeddings-based learning in the CGKN, which is analogous to the Fourier-modes-based learning in the FNO. It is worth highlighting that due to the black-box nature and lack of the DA loss constraint in the training state, the general nonlinear neural-network-based models including DNN, CNN, and FNO are incapable of performing efficient and accurate DA,  which is a key advantage of CGKN.

\begin{figure}[H]
    \centering
    \includegraphics[width=\linewidth]{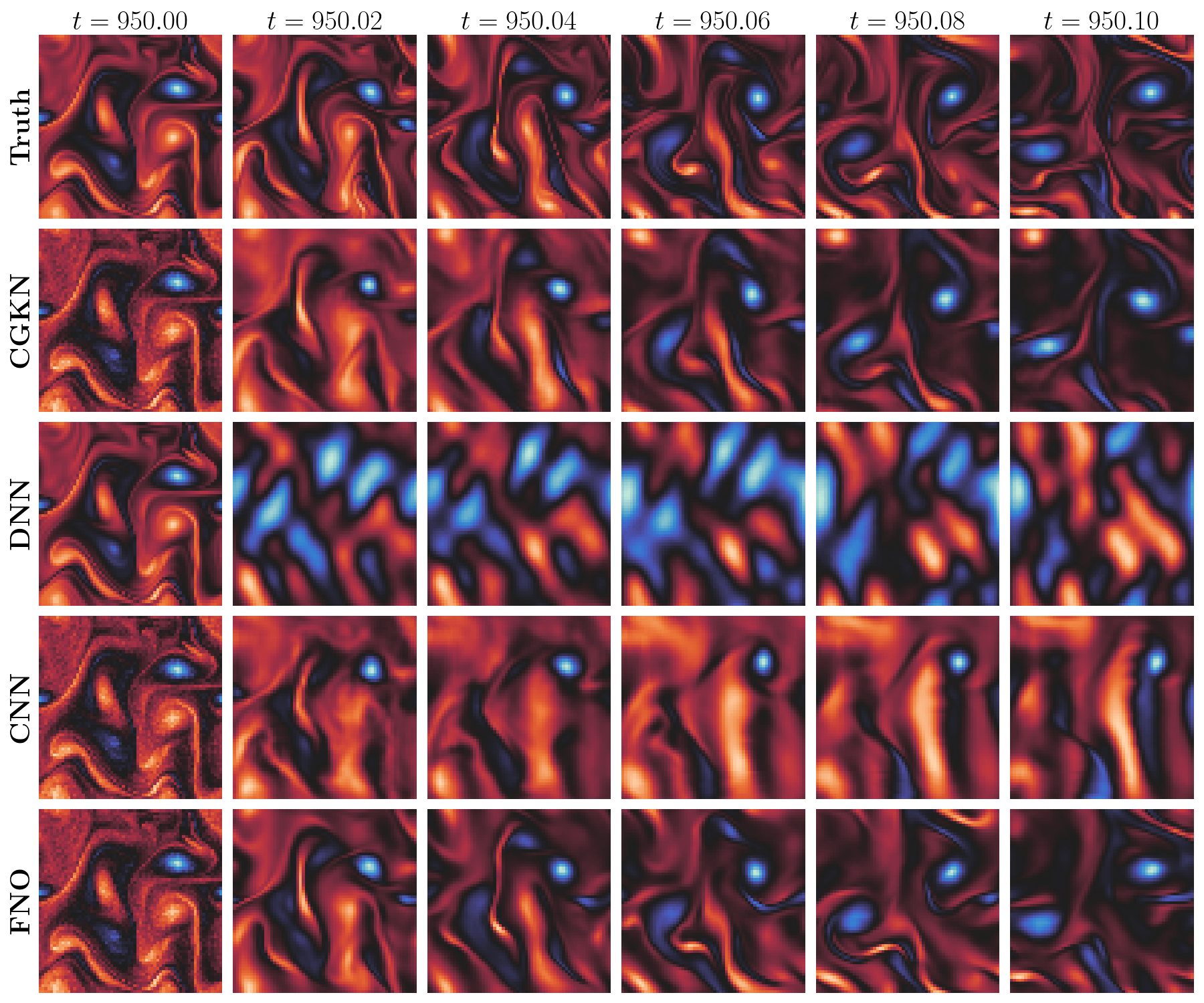}
    \caption{Results of state forecast for the Navier-Stokes equations. The evolution of the true vorticity field is compared with the predictive vorticity field from various models including CGKN, DNN, CNN, and FNO.}
    \label{fig:NSE_SF_Heatmap}
\end{figure}

The state forecast results from different models are presented in Fig.~\ref{fig:NSE_SF_Heatmap}, with the initial condition set as the vorticity at 950 time units from the test dataset. The forecast results from CGKN and FNO successfully capture the flow pattern of the true simulation, whereas the result from CNN deviates significantly, and that from DNN performs the worst. In addition to its superior state forecast performance, a major advantage of CGKN is its capability for efficient and accurate DA.

\begin{figure}[H]
    \centering
    \includegraphics[width=\linewidth]{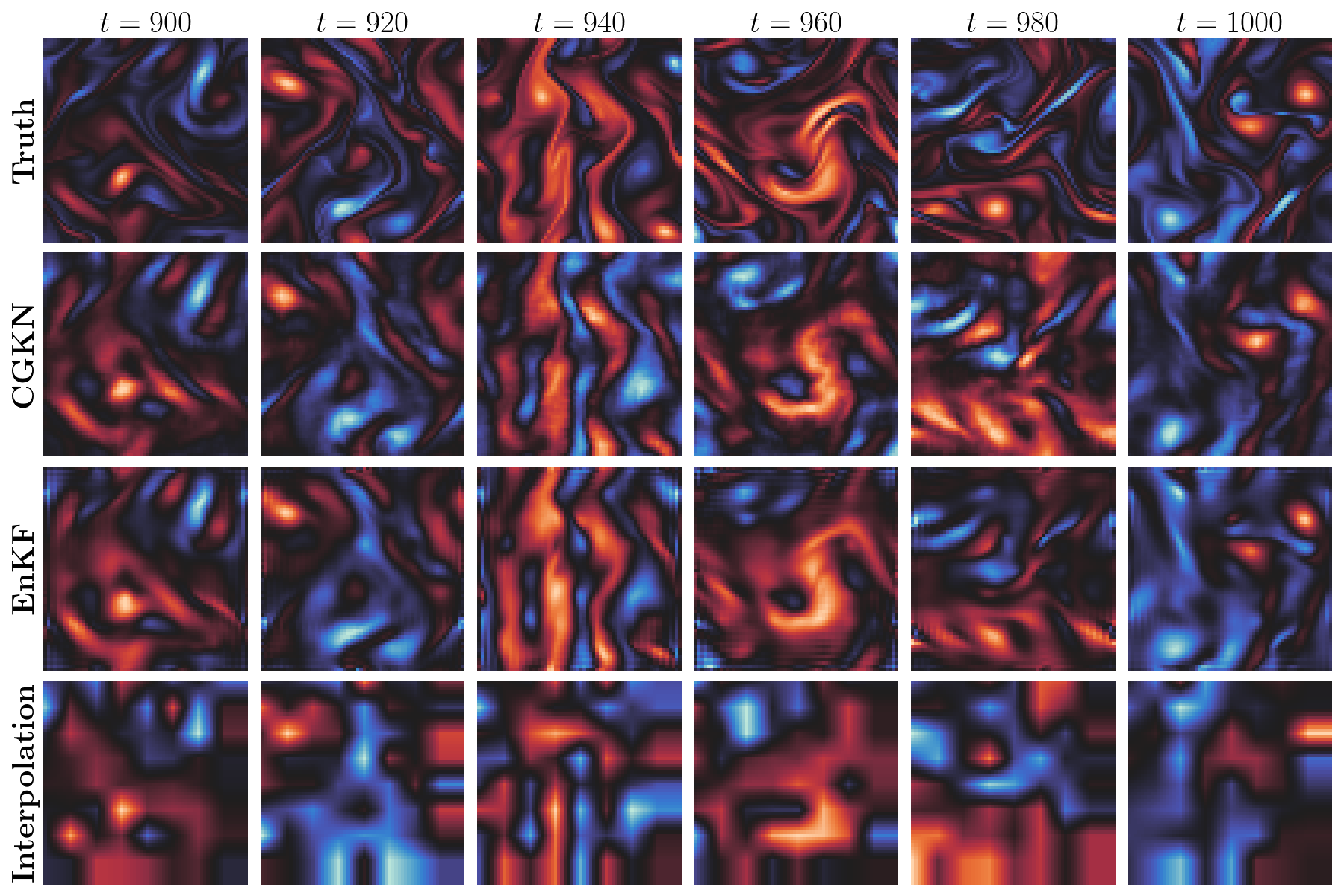}
    \caption{Spatiotemporal evolution of the DA results for the Navier-Stokes equations. Heatmaps of each row display the vorticity fields for the true simulation, the DA posterior means from CGKN and EnKF, and the interpolation result.}
    \label{fig:NSE_DA_Heatmap}
\end{figure}

Figure~\ref{fig:NSE_DA_Heatmap} shows the DA results from CGKN and EnKF. Given a trajectory of $8\times 8$ observed states uniformly distributed on the vorticity field starting from 800 time units in test data, the evolutions of the full fields ($64 \times 64$) of the true simulation, posterior means from CGKN and EnKF, and interpolation result are shown in each row of the figure. The flow patterns of estimated fields from CGKN and EnKF can capture that of the true fields, while the flow pattern of interpolation is very blurry and deviates significantly from the truth. The figure confirms the comparable DA result between CGKN and EnKF, despite CGKN using much fewer computing resources.

\begin{figure}[H]
    \centering
    \includegraphics[width=0.9\linewidth]{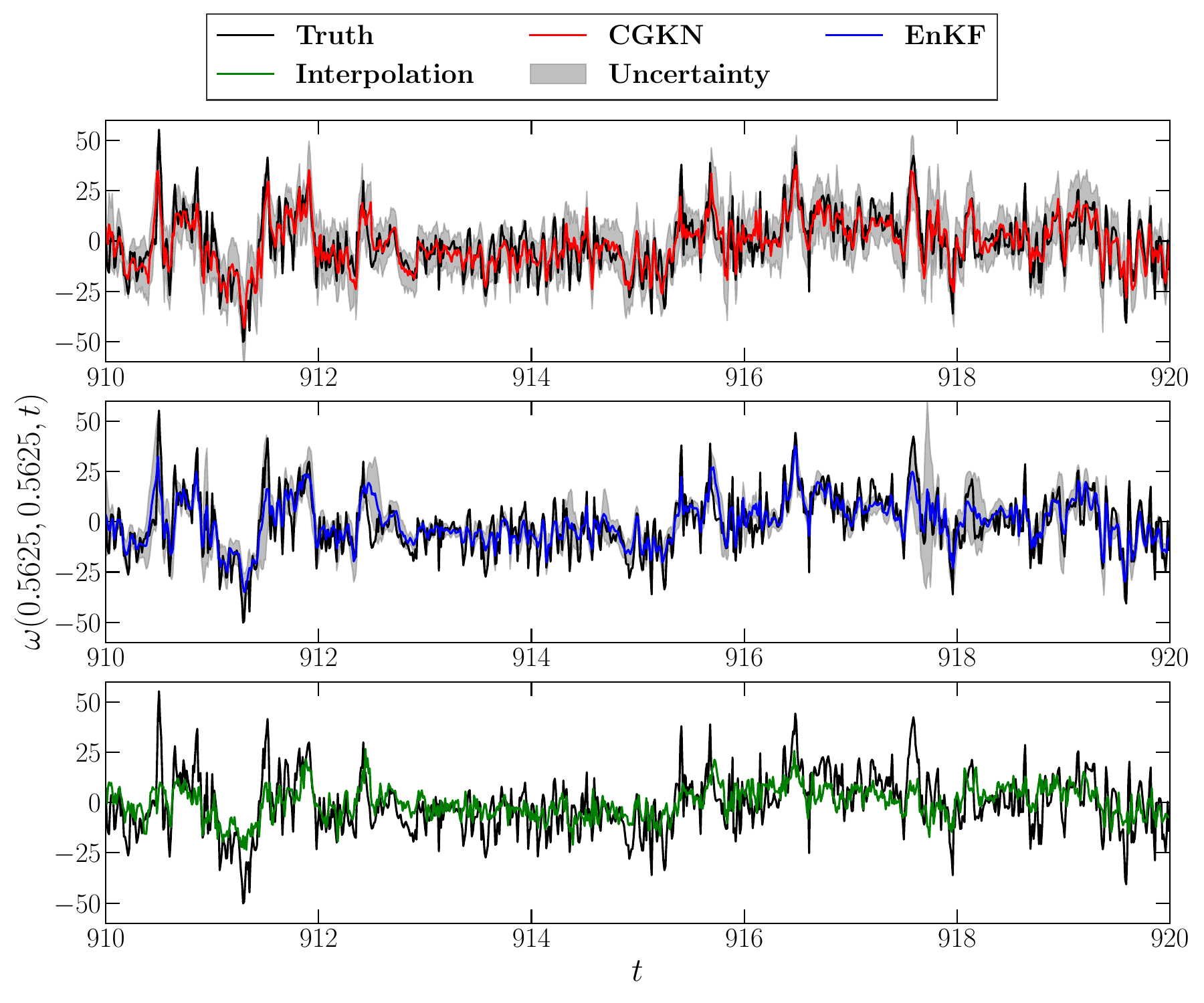}
    \caption{Time series of the DA results for the Navier-Stokes equations. The true signal, DA posterior means from CGKN and EnKF together with uncertainty areas of two standard deviations, and interpolation result for the unobserved state at spatial position $(0.5625, 0.5625)$ are presented.}
    \label{fig:NSE_DA_TS}
\end{figure}

Figure~\ref{fig:NSE_DA_TS} shows the time series of the unobserved state indexed at the location $(x, y) = $ \\
$(0.5625, 0.5625)$ in the spatial domain of $[0, 1]^2$. The first two rows of the figure present a comparison between the true signal and the DA posterior means from CGKN and EnKF, along with their associated uncertainty areas. The uncertainty area is two standard deviations associated with the posterior means, which for CGKN is determined through residual analysis, and for EnKF, it is estimated using the ensemble standard deviation. The interpolation result is shown in the last row. By comparing the posterior mean and the true signal, both CGKN and EnKF outperform the interpolation result, with CGKN demonstrating slightly better performance than EnKF, particularly for some extreme values. Additionally, the uncertainty areas from both CGKN and EnKF can effectively encompass most of the true signals, demonstrating a robust quantification of the uncertainty associated with the estimated mean field. Overall, the CGKN is capable of tackling both tasks of state forecast and efficient DA for this 2-D turbulence example. It is worth noting that the CGKN can achieve comparable DA performance to that of EnKF but with significantly lower computational costs. For state forecast, the CGKN can reach a level of performance similar to that of FNO.

\section{Conclusion and Discussion}
\label{sec:Conclusion}

This work develops a unified deep learning framework, the discrete-time conditional Gaussian Koopman network (CGKN), to learn a surrogate model to perform efficient data assimilation (DA) and state forecast for spatiotemporal dynamical systems governed by nonlinear partial differential equations. Exploiting the Koopman embedding of the unobserved system states to discover a latent representation with conditional linear dynamics, the spatiotemporal dynamical system can be converted into a conditional Gaussian nonlinear system, serving as a surrogate model of the true system and facilitating analytical formulae for DA. The analytical DA formulae and the lower-dimensional embedded latent space can significantly improve the efficiency of DA, enabling the incorporation of DA performance into the training process of the surrogate model. Therefore, the proposed discrete-time CGKN framework in this work unifies the processes of training a scientific machine learning (SciML) model and tuning its DA performance. The effectiveness of the proposed framework for efficient state forecast and DA is demonstrated through several canonical problems governed by PDEs, including the viscous Burgers' equation, the Kuramoto–Sivashinsky equation, and the 2-D Navier-Stokes equations. Extensions and future directions of the proposed framework CGKN include introducing sparsity or locality in the identified latent states to employ the spatial information for further accelerating the speed of state forecast and DA, exploring alternative approaches to quantifying the uncertainties associated with the posterior mean of CGKN, and employing advanced multi-objective optimization methods to minimize each loss functions for enhancing the training performance of CGKN. Incorporating some known physics and designing more structures of CGKN, e.g., allowing temporal memories to account for non-Markovian dynamics and characterizing the stochastic terms via fluctuation–dissipation theorem, are also interesting future directions. It is worth noting that the proposed CGKN framework also serves as an example to illustrate unifying the development of SciML models and their outer-loop applications, which can potentially inspire research directions of other outer-loop applications such as design optimization, inverse problems, and optimal control.

\section*{Acknowledgments}
The research of C.C. and J.W. was funded by the University of Wisconsin-Madison, Office of the Vice Chancellor for Research and Graduate Education, with funding from the Wisconsin Alumni Research Foundation. The research of N.C. is funded by Office of Naval Research N00014-24-1-2244 and Army Research Office W911NF-23-1-0118. Z.W. is partially supported as a research assistant under the first grant.

\section*{Data Availability}
The data that support the findings of this study are available from the corresponding author upon reasonable request.

\section*{Code Availability}
The code that supports the findings of this study is available in the link: \url{https://github.com/AIMS-Madison/Discrete-CGKN}.

\bibliographystyle{unsrt}
\bibliography{references}

\clearpage

\appendix

\section{Koopman Theory}
\label{app:Koop}

For nonlinear dynamical systems in the general form:
\begin{equation}
\label{eq:CTCS_App}
    \mathbf{u}^{n+1} = \mathcal{G}(\mathbf{u}^n),
\end{equation}
where $\mathbf{u}^n \in \mathbb{R}^d$ is the system state and $\mathcal{G}$ is a nonlinear map, Koopman theory~\citep{koopman1931hamiltonian} provides a linear perspective by describing the system in terms of the evolution of observable of the system state. The observable $h: \mathbb{R}^{d} \mapsto \mathbb{R}$ is a function in the Hilbert space operating on the system state $\mathbf{u}$.

The Koopman operator $\mathcal{K}$ is a linear operator acting on observable function $h$ which is defined as
\begin{equation}
    \mathcal{K}h := h \circ \mathcal{G},
\end{equation} where $\circ$ denotes function composition.

The evolution of observable $h$ is governed by the Koopman operator:
\begin{equation}
    h(\mathbf{u}^{n+1}) = h\circ\mathcal{G}(\mathbf{u}^n) = \mathcal{K} h(\mathbf{u}^n).
\end{equation}

In practice, this infinite-dimensional linear dynamical system and the linear operator $\mathcal{K}$ can be approximated by finite-dimensional representations:
\begin{equation}
\label{eq:LDS_App}
    \mathbf{v}^{n+1} = \mathbf{A}\mathbf{v}^n + \boldsymbol{\sigma}\boldsymbol{\epsilon}^n,
\end{equation} where $\mathbf{v}^n \in \mathbb{R}^{d_{\mathbf{v}}}$ is a finite-dimensional vector representing the linear embeddings of the system states $\mathbf{u}^n$, approximating the infinite-dimensional function $h(\mathbf{u}^n)$. The matrix $\mathbf{A} \in \mathbb{R}^{d_{\mathbf{v}} \times d_{\mathbf{v}}}$ is the linear dynamics of this discrete linear dynamical system, which approximates the linear operator $\mathcal{K}$. The $\boldsymbol{\epsilon}^n \in \mathbb{R}^{d_{\mathbf{v}}}$ is Gaussian white noise with $\boldsymbol{\sigma} \in \mathbb{R}^{d_{\mathbf{v}} \times d_{\mathbf{v}}}$ as the noise coefficient.

\section{Generalized Application of Koopman Theory for Partially Observed Systems}
\label{app:GenKoop}

Considering a partially observed nonlinear dynamical system in the general form of
\begin{equation}
\label{eq:DTDS_u1u2_App}
\begin{aligned}
    \mathbf{u}_1^{n+1} &= \mathcal{G}_1\big( \mathbf{u}_1^{n}, \mathbf{u}_2^{n}\big), \\
    \mathbf{u}_2^{n+1} &= \mathcal{G}_2\big( \mathbf{u}_1^{n}, \mathbf{u}_2^{n}\big),
\end{aligned}
\end{equation}
where $\mathbf{u}^n = \{\mathbf{u}_1^n, \mathbf{u}_2^n \}$ with observed states $\mathbf{u}^n_1 \in \mathbb{R}^{d_1}$ and unobserved states $\mathbf{u}^n_2 \in \mathbb{R}^{d_2}$. 

The generalized application of Koopman theory to the partially observed nonlinear dynamical system in Eq.~\eqref{eq:DTDS_u1u2_App} aims to linearize the unobserved states $\mathbf{u}_2^n$ while retaining the nonlinearity of the observed states $\mathbf{u}_1^n$. Therefore, the full system remains highly nonlinear, fundamentally contrasting with the standard Koopman theory. The approximated finite-dimensional dynamical system is given by: 
\begin{equation}
\begin{aligned}
\label{eq:CGNS_App}
\mathbf{u}_1^{n+1} &= \mathbf{F}_1\big(\mathbf{u}_1^{n}\big) + \mathbf{G}_1\big(\mathbf{u}_1^{n}\big)\mathbf{v}^{n} + \boldsymbol{\sigma}_1\boldsymbol{\epsilon}_1^{n},\\
\mathbf{v}^{n+1} &= \mathbf{F}_2\big(\mathbf{u}_1^{n}\big) + \mathbf{G}_2\big(\mathbf{u}_1^{n}\big)\mathbf{v}^{n} + \boldsymbol{\sigma}_2\boldsymbol{\epsilon}_2^{n},
\end{aligned}
\end{equation}
where $\mathbf{v}^n \in \mathbb{R}^{d_{\mathbf{v}}}$ are linear embeddings of unobserved states $\mathbf{u}_2^n$ and $\mathbf{F}_1$, $\mathbf{G}_1$, $\mathbf{F}_2$, $\mathbf{G}_2$ are four nonlinear maps of observed states $\mathbf{u}_1^n$. The $\boldsymbol{\epsilon}_1^n \in \mathbb{R}^{d_1}$ and $\boldsymbol{\epsilon}_2^n \in \mathbb{R}^{d_{\mathbf{v}}}$ are independent Gaussian white noises with $\boldsymbol{\sigma}_1 \in \mathbb{R}^{d_1 \times d_1}$ and $\boldsymbol{\sigma}_2 \in \mathbb{R}^{d_{\mathbf{v}} \times d_{\mathbf{v}}}$ as noise coefficients.

\section{Conditional Gaussian Nonlinear Systems}
\label{app:CGNS}

The system in Eq.~\eqref{eq:CGNS_App} belongs to a class of nonlinear stochastic dynamical systems called conditional Gaussian nonlinear systems (CGNS). A key feature of the CGNS is that the conditional distribution $p(\mathbf{v}^{n} | \{\mathbf{u}_1^{i}\}_{i=0}^{n})$ is Gaussian, whose mean $\boldsymbol{\mu}_{\mathbf{v}}$ and covariance $\boldsymbol{\Sigma}_{\mathbf{v}}$ can be solved by the analytical formulae \citep{liptser2013statistics}:
\begin{equation}
\label{eq:CGFilter_App}
    \begin{aligned}
    \boldsymbol{\mu}_{\mathbf{v}}^{n+1} &= \mathbf{F}_2 + \mathbf{G}_2\boldsymbol{\mu}_{\mathbf{v}}^{n} + \mathbf{K}^{n} \big(\mathbf{u}_1^{n+1} - \mathbf{F}_1 - \mathbf{G}_1\boldsymbol{\mu}_{\mathbf{v}}^{n}\big),\\
    \mathbf{\boldsymbol{\Sigma}}_{\mathbf{v}}^{n+1} &= \mathbf{G}_2\mathbf{\boldsymbol{\Sigma}}_{\mathbf{v}}^{n}\mathbf{G}_2^\mathtt{T} + \boldsymbol{\sigma}_2\boldsymbol{\sigma}_2^\mathtt{T} - \mathbf{K}^{n}\mathbf{G}_1\mathbf{\boldsymbol{\Sigma}}^{n}_{\mathbf{v}}\mathbf{G}_2^\mathtt{T},
    \end{aligned}
\end{equation}
where $\mathbf{K}^{n} = \mathbf{G}_2 \mathbf{\boldsymbol{\Sigma}}_{\mathbf{v}}^{n}\mathbf{G}_1^\mathtt{T}\big(\boldsymbol{\sigma}_1\boldsymbol{\sigma}_1^\mathtt{T} + \mathbf{G}_1\mathbf{\boldsymbol{\Sigma}}_{\mathbf{v}}^{n}\mathbf{G}_1^\mathtt{T}\big)^{-1}$ is an analog to the Kalman gain. The conditional distribution is the filtering solution of data assimilation and the formulae in Eq.~\eqref{eq:CGFilter_App} is called the conditional Gaussian filter (CG Filter).

\end{document}